%% file: main.tex
\documentclass[runningheads]{llncs}

\usepackage{eccv}

\usepackage{eccvabbrv}

\usepackage{graphicx}
\usepackage{booktabs}

\usepackage[pagebackref,breaklinks,colorlinks,citecolor=eccvblue]{hyperref}

\usepackage[dvipsnames]{xcolor}

\newcommand\ul[1]{\underline{#1}}
\newcommand{\RE}[1]{{\color{red}#1}}
\newcommand{\BL}[1]{{\color{blue}#1}}
\definecolor{newgreen}{rgb}{0, 0.6, 0.2}

\usepackage{array}
\usepackage{multirow}
\usepackage{colortbl}
\usepackage{arydshln} %
\usepackage{svg}  %
\usepackage{amsmath}
\usepackage{tabularx}
\usepackage{makecell}
\usepackage{threeparttable}  %

\begin{document}

\title{LightAvatar: Efficient Head Avatar as Dynamic Neural Light Field}

\titlerunning{LightAvatar: Efficient Head Avatar as Dynamic NeLF}

\author{Huan Wang\inst{1,2,\dag}\and
Feitong Tan\inst{2} \and
Ziqian Bai\inst{2,3} \and
Yinda Zhang\inst{2} \and
Shichen Liu\inst{2} \and
Qiangeng Xu\inst{2} \and
Menglei Chai\inst{2} \and
Anish Prabhu\inst{2} \and
Rohit Pandey\inst{2} \and
Sean Fanello\inst{2} \and
Zeng Huang\inst{2} \and
Yun Fu\inst{1}}
\authorrunning{Huan Wang, et al.}
\institute{Northeastern University, USA \and Google, USA \and Simon Fraser University, Canada}
\renewcommand{\thefootnote}{\fnsymbol{footnote}}
\footnotetext{$^\dag$Work done when Huan was an intern at Google.}
\footnotetext{Corresponding author: Huan Wang, \tt{huan.wang.cool@gmail.com}.}

\maketitle
\input{sec/teaser_fig}  
\input{sec/0_abstract}    
\input{sec/1_intro_v2}

\input{sec/2_related_work}
\input{sec/3_method}

\input{sec/4_experiments}

\input{sec/5_conclusion}

\bibliographystyle{splncs04}
\bibliography{main}

\newpage
\input{sec/6_supp_eccv24_workshop}
\end{document}

%% file: sec/teaser_fig.tex
\vspace{-3mm}
\begin{figure}
\centering
\resizebox{0.995\linewidth}{!}{
\begin{tabular}{c@{\hspace{0.03\linewidth}}c@{\hspace{0.01\linewidth}}c}
\includegraphics[width=0.55\linewidth]{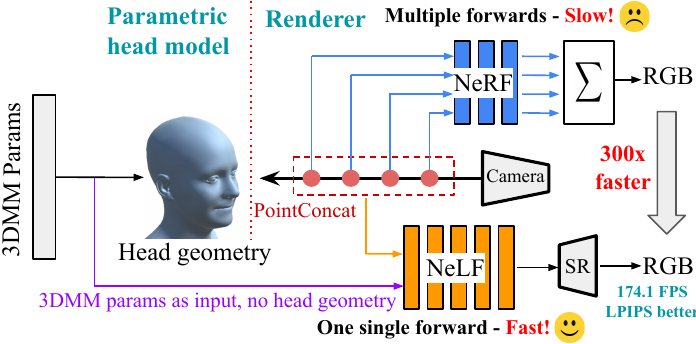} &
\includegraphics[width=0.38\linewidth]{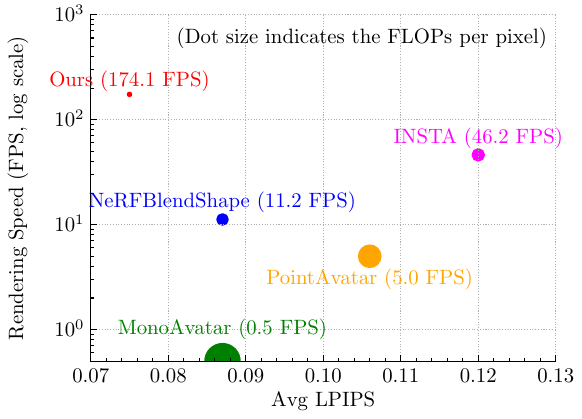} \\
\small (a) NeRF-based Avatars~\vs~our LightAvatar & 
\small (b) FPS-LPIPS comparison
\end{tabular}}
\vspace{-2mm}
\caption{(a) Overview comparison between existing neural head avatars (top)~\vs~our LightAvatar (down) -- a brand-new framework to build efficient 3D head avatars based on neural light field. LightAvatar features a simple and uniform design, which takes expression code and camera pose as input, renders the RGB via a single network forward pass, running at \textbf{174.1 FPS} (on a RTX3090 GPU) with image quality improved.
(b) FPS and LPIPS comparison of recent top-performing (fast) avatars. Our method achieves much faster rendering speed with better LPIPS than the counterparts.
}
\label{fig:teaser}
\vspace{-12mm}
\end{figure}

%% file: sec/0_abstract.tex
\begin{abstract}
Recent works have shown that neural radiance fields (NeRFs) on top of parametric models have reached SOTA quality to build photo-realistic head avatars from a monocular video.
However, one major limitation of the NeRF-based avatars is the slow rendering speed due to the dense point sampling of NeRF, preventing them from broader utility on resource-constrained devices. We introduce \textbf{LightAvatar}, the \textit{first} head avatar model based on neural \textit{light} fields (NeLFs). LightAvatar renders an image from 3DMM parameters and a camera pose via a single network forward pass, without using mesh or volume rendering.
The proposed approach, while being conceptually appealing, poses a significant challenge towards real-time efficiency and training stability. To resolve them, we introduce dedicated network designs to obtain proper representations for the NeLF model and maintain a low FLOPs budget. Meanwhile, we tap into a distillation-based training strategy that uses a pretrained avatar model as teacher to synthesize abundant pseudo data for training. A warping field network is introduced to correct the fitting error in the real data so that the model can learn better. 
Extensive experiments suggest that our method can achieve new SOTA image quality quantitatively or qualitatively, while being significantly faster than the counterparts, reporting \textbf{174.1 FPS} (512$\times$512 resolution) on a consumer-grade GPU (RTX3090) with no customized optimization.
\end{abstract}

%% file: sec/1_intro_v2.tex
\section{Introduction}
\label{sec:intro}
Digitalizing a human is a long-standing problem in computer vision~\cite{blanz1999morphable}, which has recently attracted increasing attention due to its massive potential in AR/VR. 
One of the most prominent task is to build face or head avatars, which can generate vivid and photo-realistic appearance of the users upon controlling signals, \eg, from parametric models~\cite{li2017learning,egger20203d}.
Recent advances in implicit neural representation (INR)~\cite{park2019deepsdf,mescheder2019occupancy,chen2019learning} has motivated neural radiance field (NeRF)~\cite{mildenhall2020nerf}, which has been demonstrated as an enormous success in building controllable photo-realistic 3D head avatars~\cite{gafni2021dynamic,zheng2022avatar,bai2023learning}.

Despite the remarkable rendering quality, existing NeRF-based avatars typically suffers from two issues.
First, the rendering is computationally expensive, which is an issue arising from the neural volumetric rendering backbone due to multiple shading operations along each pixel ray.
Although quite many works have attempted to accelerate \textit{static} NeRFs~\cite{reiser2021kilonerf,rebain2021derf,garbin2021fastnerf,yu2021plenoctrees,wang2022r2l,cao2023real,chen2023mobilenerf}, it is non-trivial to extend them to \textit{dynamic} head avatars.
Second, most of the high-quality avatars build upon explicit 3DMM~\cite{blanz1999morphable,egger20203d} geometry, \eg, to derive a 3D deformation field~\cite{grassal2022neural,athar2022rignerf,zielonka2023instant}, or directly anchor local radiance fields~\cite{bai2023learning}.
While delivering remarkable controllability and stability, these methods tend to perform poor when 3DMM geometry is over-simplified or missing.

In this paper, we present \textit{LightAvatar}, a novel 3D head avatar model that renders high-quality images efficiently without leveraging an explicit geometry.
Our model is fundamentally different from existing approaches in the rendering backbone (see Fig.~\ref{fig:method_overview}), where we leverage a neural \textit{light} field (NeLF) instead of a neural \textit{radiance} field (NeRF) as scene representation.
The rendering of light fields amounts to \textit{a single} neural network forward pass \textit{vs.}~hundreds of network forward passes in NeRF~\cite{mildenhall2020nerf}. The network efficiently renders a target image by a single forward pass, with 3DMM parameters and camera pose as input and transformed by our dedicatedly designed sub-networks into better representations for the NeLF backbone.
To further speed up the rendering, we introduce an image \textit{super-resolution (SR)}~\cite{yang2019deep,lim2017enhanced,zhang2018image} module after the NeLF backbone in our pipeline, which enables us to feed a low-resolution input into the network while obtain a high-resolution output.  
As a result, the rendering is substantially faster than the other counterparts (174.1 FPS, see Tab.~\ref{tab:speed_comparison}).

One downside of removing the dependency on explicit 3DMM geometry, however, is less training stability due to the missing of strong priors, especially when training from a monocular video.
To overcome this, we tap into the recent advances of \textit{knowledge distillation}~\cite{bucilua2006model,hinton2015distilling} in efficient neural rendering~\cite{wang2022r2l,cao2023real,yu2023dylin}. Specifically, we employ a pretrained avatar model to synthesize abundant pseudo data, and distill a LightAvatar model from them. 
To prevent the performance from being capped by the teacher model, we train jointly on both the pseudo and real data.
While perfect 3DMM fitting is guaranteed on the pseudo data, this is not true on the real data.
We observed that naively adding real data may even hurt the performance.
To account for the fitting error in the real data, we introduce a \textit{warping field network} to mitigate the fitting error in the real data, resulting in improved overall quality. Contributions of this work are:
\begin{itemize}
    \item We introduce \textit{LightAvatar}, the \textit{first} head avatar based on neural light fields (NeLFs) that does not rely on explicit meshes or volume rendering. This novel approach results in a simple and efficient pipeline.
    \item The method features several dedicated network designs: (1) The expression representation produced by our model outperforms the common baseline solution of using raw expressions as NeLF model input. (2) Importantly, we introduce an SR module that significantly improves the inference efficiency.
    \item We present a distillation-based training strategy with a warping field network for correcting fitting error to learn effectively on pseudo data and real data. 
    \item Extensive empirical results and analyses on multiple subjects show our LightAvatar can achieve \textit{consistently better} image quality (see Tab.~\ref{tab:results_quantitative_flame}, Tab.~\ref{tab:quantitative_results_fast_avatars}, Fig.~\ref{fig:results_qualitative_flame}, Fig.~\ref{fig:results_qualitative_fast_avatars}) than the counterparts while running at \textbf{174.1 FPS} on a RTX3090 GPU with no customized optimization (see Tab.~\ref{tab:speed_comparison}).
\end{itemize}

%% file: sec/2_related_work.tex
\section{Related Work}
\label{sec:related_work}
\subsection{Monocular 3D Head Avatars and Fast Avatars}
\noindent \textbf{Monocular 3D Head Avatars}. It has been a long challenge to reconstruct 3D head avatars from a monocular video. 
Leveraging the low-dimensional priors when modeling human heads (\eg, 3DMM~\cite{blanz1999morphable,egger20203d}), many previous works model the geometry and texture \textit{explicitly}, such as~\cite{weise2011realtime,garrido2014automatic,ichim2015dynamic,thies2015real,cao2016real,garrido2016reconstruction,thies2016face,hu2017avatar,kim2018deep,thies2020neural}. 
By fitting a morphable model to a given subject through traditional optimization~\cite{zollhfer2018state} or neural-based techniques~\cite{tewari2019fml,bai2020deep,chaudhuri2020personalized,yang2020facescape,bai2021riggable}, they employ shared mesh topology and texture parameterization, enabling subsequent animation and manipulation. 
Nevertheless, these approaches often face challenges due to their restricted representation ability, especially in modeling subtle details and components such as hair and accessories that fall outside the model parameters.

\begin{figure*}[t]
\centering
\begin{tabular}{c}
\includegraphics[width=1.0\linewidth]{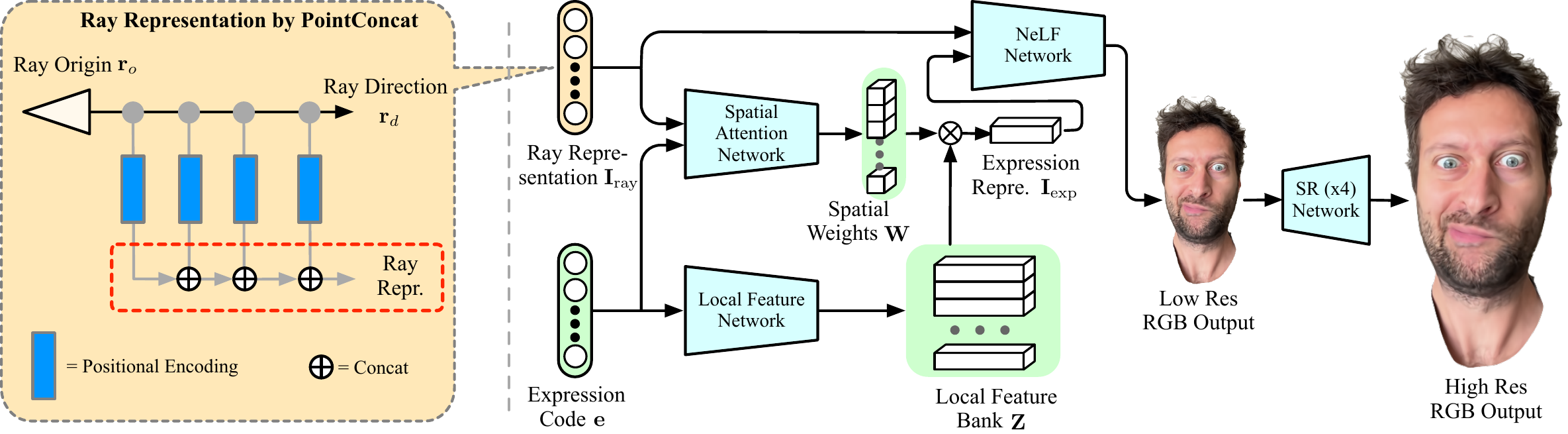}
\end{tabular}
\vspace{-2mm}
\caption{Overview of our LightAvatar. The method consists of four trainable networks (\textit{spatial attention network}, \textit{local feature network}, \textit{NeLF network}, and \textit{SR network}). \textbf{(1)} Given an expression code, the local feature network transforms it to a \textit{local feature bank}, which stores the features of different local head regions. \textbf{(2)} Given a specific ray and the expression code, the spatial attention network outputs a vector of \textit{spatial attention weights} to query the local feature bank to obtain the expression representation for that ray. \textbf{(3)} Then, given the ray and expression representation as input, the NeLF model predicts the desired (low-resolution) RGB. \textbf{(4)} Finally, the SR network generates a high-resolution image with the low-resolution image as input. Notably, LightAvatar predicts the target RGB via a single network forward, thus enabling fast rendering.
}
\label{fig:method_overview}
\vspace{-3mm}
\end{figure*}

The rapid advancement of implicit neural representations (INR), represented by NeRF~\cite{mildenhall2020nerf,barron2021mip,barron2022mip} has recently popularized the implicit modeling of avatars, thanks to its superior rendering quality and the ability to comprehensively represent the entire head. 
For instance, NerFACE~\cite{gafni2021dynamic} directly conditions neural radiance fields with 3DMM expression codes to achieve dynamic avatars. 
RigNeRF~\cite{athar2022rignerf}, on the other hand, employs expression conditioning in the canonical space, defined by a 3DMM-guided warping field. 
IMAvatar~\cite{zheng2022avatar} endeavors to learn blendshapes and skinning fields to represent deformations dependent on avatar expressions and poses. 
Additionally, MonoAvatar~\cite{bai2023learning} focuses on learning local features anchored on the 3DMM geometry, allowing these features to be deformed by 3DMM animation and interpolated within the 3D volume to facilitate avatar animation.
Despite the encouraging progress, a primary challenge remains in achieving precise control over motions and expressions, and improving the efficiency during training and inference~\cite{ma2021pixel,zheng2023pointavatar,zielonka2023instant}.

\vspace{0.5em}
\noindent \textbf{Fast Avatars}. Many recent papers have tried to improve the efficiency of head avatars. 
INSTA~\cite{zielonka2023instant} uses neural graphics primitives~\cite{muller2022instant} embedded on a parametric face model, achieving fast reconstruction in less than 10 minutes and interactive rendering.
NeRFBlendShape~\cite{gao2022reconstructing} adopts multi-level voxel fields as the bases citing the idea of Instant-NGP~\cite{muller2022instant}, where each voxel is modeled by a lightweight NeRF, enabling fast training and rendering.
FLARE~\cite{bharadwaj2023flare} presents a relightable mesh-based avatar where materials and lighting are disentangled by different MLPs. Hash-grid encoding~\cite{muller2022instant}, neural split-sum approximation~\cite{karis2013real}, and differentiable rasterization~\cite{laine2020modular} are used for fast training and rendering.
Using neither mesh nor neural implicit representations, PointAvatar~\cite{zheng2023pointavatar} employs a deformable point-cloud-based representation, which can be rendered efficiently with standard differentiable rasterizer.
Despite their improved rendering efficiency, the visual quality of most of these methods is yet to be satisfactory according to our empirical studies (Fig.~\ref{fig:results_qualitative_fast_avatars}). In this work, we aim to deliver an avatar of both fast rendering speed \textit{and} high quality.

\subsection{Efficient Neural Rendering}
\textbf{NeRF Inference Acceleration}. While neural rendering methods~\cite{tewari2022advances}, particularly NeRFs~\cite{mildenhall2020nerf,barron2021mip,barron2022mip}, offer superior quality in synthesizing novel views, their computational demands during inference often pose challenges for real-time scenarios and resource-constrained devices. Efforts to enhance the efficiency of general NeRF model rendering have primarily pursued three directions. Some methods~\cite{garbin2021fastnerf,hedman2021baking,yu2021plenoctrees,sun2022direct} optimize the rendering by employing precomputation, shifting from MLP forward passes to table lookup. Other approaches ~\cite{rebain2021derf,reiser2021kilonerf,chen2022tensorf,muller2022instant} remap the entire NeRF scene into more efficient spatial structures that can be parallelized, achieving consistent speedups. Finally, the rendering speed can be boosted by reducing the number of samples per camera ray~\cite{lindell2021autoint,neff2021donerf,chen2023mobilenerf}, either within the original NeRF architecture or by embracing alternative frameworks such as 
neural light fields (NeLFs)~\cite{wang2022r2l,liu2022neulf,cao2023real,gupta2023lightspeed}.

\vspace{0.5em}
\noindent\textbf{Neural Light Fields} (NeLFs) have emerged as a promising solution for rapid neural rendering thanks to their distinct advantage of requiring only a single forward pass, in contrast to numerous samples per ray in NeRF. While light fields have been studied~\cite{levoy1996light,gortler1996lumigraph} as efficient scene representations in real-time image-based rendering in computer graphics, their recent adaptation to neural networks~\cite{kalantari2016learning,mildenhall2019local,bemana2020x} enables efficient implicit scene modeling. Among them, light field networks~\cite{sitzmann2021light}, light field neural rendering~\cite{suhail2022light}, RSEN~\cite{attal2022learning}, NeuLF~\cite{liu2022neulf}, R2L~\cite{wang2022r2l}, MobileR2L~\cite{cao2023real}, and LightSpeed~\cite{gupta2023lightspeed} have proposed various strategies for efficient NeLF-based scene representations.
Our work aligns with this trend by modeling neural avatars as NeLF instead of NeRF to significantly accelerate inference. However, compared to existing methods that primarily address static scenes, training dynamic NeLF on animatable avatars presents greater challenges, partially due to the high demand of training data, specifically in terms of scale, diversity, and the precision required for alignment.

%% file: sec/3_method.tex
\section{Proposed Method: LightAvatar} \label{sec:method}
\subsection{Prerequisites: NeRF-based Avatars and NeLF}
A head avatar typically has two major components during inference:  \textit{a parametric head model} and \textit{renderer}.
The parametric model describes the head geometry under different expressions.
The renderer is responsible to synthesize the final head images, using the supervision of RGB data from a monocular video. 

In this paper, FLAME~\cite{li2017learning} is adopted as the parametric model due to its extensive usage (notably, our method can be seamlessly generalized to other parametric models).
For rendering, many top-performing implicit head avatars (\eg, NerFACE~\cite{gafni2021dynamic}, MonoAvatar~\cite{bai2023learning}) employ a NeRF-based representation that maps a 5D input (point position and viewing direction) to a 4D output (RGB and density). In inference, multiple points along a ray are sampled. For each point, the NeRF network is queried to obtain its RGB and density. Finally, the color of each ray is obtained via alpha compositing~\cite{kajiya1984ray,max1995optical,mildenhall2020nerf}.

One major problem preventing fast inference in NeRF and also NeRF-based avatars is that the the number of sampled points is pretty large. Consequently, the rendering computation for even a single pixel is prohibitively heavy, limiting the NeRF-based avatars usage in real-time applications. 

\vspace{0.5em}
\noindent \textbf{Neural Light Field (NeLF)}. This paper attempts at solving the aforementioned slow rendering problem by introducing a \textit{light field based} avatar. Unlike NeRF, NeLF learns a mapping from a camera ray to 3D RGB directly,  $\mathcal{F}_{\pmb{\Theta}}: \mathbb{R}^4 \mapsto \mathbb{R}^3$, without any alpha compositing. Rendering with NeLFs amounts to a single network forward~\vs~hundreds of network forward passes with NeRFs, thus being much faster. Although existing works~\cite{wang2022r2l,cao2023real} showed compelling NeLFs for \textit{static} scenes, to our best knowledge, there are no successfully attempts that have utilized it for building photo-realistic head avatars (which are \textit{dynamic}). This work is meant to bridge this gap.

\subsection{LightAvatar: A Dynamic NeLF-based Avatar} \label{subsec:lightavatar}
As illustrated in Fig.~\ref{fig:method_overview}, our LightAvatar has two inputs, \textit{a ray representation} and \textit{an expression representation}, and directly predicts the RGB of the ray, without explicitly relying on any geometry. The rendering process is equivalent to a single network forward.
The ray representation encodes ray position information; expression representation encodes \textit{view-dependent} expression information. In the following, we detail our design choices for both these representations.

\vspace{0.5em}
\noindent \textbf{(1) Ray Representation}. Following prior NeLF works~\cite{wang2022r2l,cao2023real}, we use the \textit{PointConcat} scheme to obtain the ray representation. Specifically, given a camera ray (its origin $\pmb{r}_o$ and direction $\pmb{r}_d$), we sample $K$ evenly-spaced points along the ray between near plane and far plane. Then, these point coordinates are concatenated as a long vector,
\begin{equation}
\small
    \pmb{I}_{\text{ray}} = (x_1, y_1, z_1, x_2, y_2, z_2, ..., x_K, y_K, z_K).
\label{eq:ray_repre}
\end{equation}
Following prior works~\cite{mildenhall2020nerf,wang2022r2l,cao2023real}, the coordinates in the ray representation are further transformed by positional encoding~\cite{vaswani2017attention} to enrich their expressive power.

\vspace{0.5em}
\noindent \textbf{(2) Expression Representation}. Differently from the rays, that capture a specific local region, the expression code itself is a global descriptor. Therefore, for different rays, the corresponding expression representation should be discriminative, depending on the specific viewing direction. Based on these considerations, the expression representation is designed to be made up of two parts, \textit{a local feature bank} and \textit{spatial attention weights}.

The local feature bank stores local features $\pmb{Z} \in \mathbb{R}^{N_\mathrm{lf} \times D_\mathrm{lf}}$ of different head regions (such as eyes, nose, mouth, \etc.). It is obtained via an MLP called \textit{local feature network} (see Fig.~\ref{fig:method_overview}) from the expression code.

The spatial attention weights are a vector $\pmb{W} \in \mathbb{R}^{1 \times N_\mathrm{lf}}$, which is the output of an MLP network called \textit{spatial attention network} (see Fig.~\ref{fig:method_overview}). Given a specific ray and an expression code, the spatial attention network generates weights that indicate the degree of attention to be allocated to distinct spatial regions. For instance, if the ray intersects an eye region, the spatial attention network should prioritize local features of that region. This approach makes the expression representation more view-dependent. Given the local feature bank $\pmb{Z}$ and spatial attention weights $\pmb{W}$, the final expression representation for a ray is obtained via the following \textit{matrix multiplication},
\begin{equation}
\pmb{I}_{\text{exp}} = \pmb{W} \cdot \pmb{Z}.
\label{eq:expressoion_repre}
\end{equation}

\noindent \textbf{NeLF Model Input}. The combination of the ray representation (Eq.~\eqref{eq:ray_repre}) and the view-dependent expression representation (Eq.~\eqref{eq:expressoion_repre}) constitutes the input of the NeLF model $\mathcal{F}_{\pmb{\Theta}}$. Specifically, they are concatenated together to form a single vector as the model input:
\begin{equation}
\small
    \pmb{I} = \mathrm{Concat}(\pmb{I}_{\text{ray}}, \pmb{I}_{\text{exp}}).
\end{equation}
Then, the RGB is predicted via a single network forward,
\begin{equation}
    \pmb{\hat{c}} = \mathcal{F}_{\pmb{\Theta}}(\pmb{I}).
\end{equation}

\noindent \textbf{SR Model for Upsampling}. To reduce the computation cost, we introduce an image super-resolution (SR) network (with $\times4$ scale), following the NeLF network (as shown in Fig.~\ref{fig:method_overview}):
\begin{equation}
    I_{SR} = \mathcal{SR}_{\pmb{\Phi}}(I_{LR}),
\end{equation}
where $I_{LR} \in \mathbb{R}^{h\times w\times 3}$, $I_{SR} \in \mathbb{R}^{h*4\times w*4\times 3}$. The $I_{SR}$ is the final output RGB.

\vspace{0.5em}
\noindent \textbf{Network Architecture}. The architectures of the NeLF backbone, spatial attention network, local feature network, and SR network are detailed as follows.

\textit{(i) NeLF backbone}. It has three parts: head, body, and tail. The head is a one-layer MLP which maps the input to the internal 128-D feature. The body consists of many residual MLP blocks of width 128 (each block has two layers). The tail is also a one-layer MLP which maps the feature to RGB. There is also a skip connection between the head and tail.

\textit{(ii) Spatial attention network and local feature network.} Similar to the NeLF model, the spatial attention network and local feature network also have three parts: head, body, and tail. The difference is that the body is much shallower, made by only 2 residual blocks with no skip connection used between the head and tail. The final output activation of spatial attention network is \textit{Sigmoid} to ensure that the outputs are in the range $(0, 1)$.

The local feature bank of the local feature network is designed as a matrix of $\mathbb{R}^{64\times 128}$, \ie, $N_\text{lf}=64, D_\text{lf}=128$. Ideally, a large local feature bank leads to more fine-grained local features. However, in practice, we do not observe significant performance improvement when using an excessively large local feature bank, while causing the cost of increased peak memory and inference time.

\textit{(iii) SR network architecture}. Similar to the NeLF backbone, the SR network also consists of three parts: head, body, and tail. The head is a single Conv layer. The body consists of multiple residual blocks (10 in this work). In the tail, there are two upsampling layers for super-resolution and another Conv layer to project back to RGB. We use \textit{Transpose Conv} layers (stride = 2) for $\times 2$ upsampling, inspired by prior work MobileR2L~\cite{cao2023real}. Yet importantly, MobileR2L~\cite{cao2023real} distributes the upsampling layers throughout the body of the SR network. This would lead to sizable increase of FLOPs because after each upsampling, the feature map size doubles, meaning the FLOPs scales by a factor of 4$\times$ thereafter. Instead, we defer the upsampling to the lightweight tail, tapping into the modern SR architecture design wisdom~\cite{lim2017enhanced}. By doing so, the major backbone of the SR network maintains a compact spatial feature map size, which is critical to the ultra-low FLOPs of our model (Tab.~\ref{tab:speed_comparison}).

\vspace{0.5em}
\noindent \textbf{Incorporating Shoulders in the Head Avatar}. 
Head and shoulders can rotate in different directions. Such a morphable structure creates extra challenges for digitizing a human. Prior related works typically tackle the head and shoulder or torso separately, \ie, using two components to render the head and shoulders respectively, while we aim at modeling the two parts with a single model.

Specifically, to capture and render the shoulders, we simply add a new set of rays representation \textit{using the shoulder rotation} tracked by off-the-shelf fitting algorithms~\cite{li2017learning}. This simply means the NeLF network input in Eq.~\eqref{eq:ray_repre} becomes a \textit{longer} vector, no more extra design needed. Essentially, this design implies that we believe the model has sufficient capacity and the input we provide has sufficient information to predict the RGB 
of both head and shoulders. %

\subsection{Training via Distillation} \label{subsec:training_via_kd}
Despite the attractive simplicity of our method, our preliminary results suggest that training the proposed LightAvatar model on the original data does not offer satisfactory performance. This problem can be solved by training the model with \textit{sufficient} data. Thus, we propose to employ a \textit{pretrained} teacher model (MonoAvatar~\cite{bai2023learning}) to synthesize more pseudo data. Specifically, given a pseudo input (expression $\pmb{e}$ and camera pose - camera origin $\pmb{r}_o$ and rotation $\pmb R$), we query the teacher model $\mathcal{T}$ to obtain the output RGB image,
\begin{equation}
    \pmb{c}^{(t)} = \mathcal{T}(\pmb{e}, \pmb{r}_o, \pmb{R}),
\end{equation}
where the expression and camera pose are obtained by \textit{interpolating} two frames randomly drawn from the real data:
\begin{equation}
\begin{aligned}
    \pmb{e} &= \alpha \cdot \pmb{e}_1 + (1 - \alpha) \cdot \pmb{e}_2, \\
    \pmb{r}_o &= \alpha \cdot \pmb{r}_{o1} + (1 - \alpha) \cdot \pmb{r}_{o2}, \\
    \pmb{R} &= \alpha \cdot \pmb{R}_{1} + (1 - \alpha) \cdot \pmb{R}_{2}, \\
\end{aligned}
\end{equation}
where $\alpha \sim \mathrm{U[0, 1)}$. The training example of pseudo data is organized as,
\begin{equation}
(\pmb{r}_o, \pmb{r}_d, \pmb{e}, \pmb{c}^{(t)}),
\label{eq:data_format}
\end{equation}
where $\pmb{r}_o, \pmb{r}_d, \pmb{e}$ represent the ray origin, direction, and expression, respectively. These together make the the input of our LightAvatar model. For the real data, we use a similar format and replace $\pmb{c}^{(t)}$ with the real captured RGB $\pmb{c}$.

\vspace{0.5em}
\noindent \textbf{Loss function}. Our LightAvatar model $\mathcal{F}_{\pmb\Theta}$ is trained with the photometric loss that minimizes the $L_2$ distance between the predicted image $I_{SR}$ and target image $I_{target}$ and a perceptual loss~\cite{johnson2016perceptual},
\begin{equation}
\mathcal{L} = || I_{SR} - I_{target} ||^2_2 + \lambda \mathcal{P}(I_{SR}, I_{target}),
\label{eq:loss}
\end{equation}
where $I_{\text{target}}$ can come from pseudo image or real image; $\mathcal{P}$ means perceptual loss with VGG network~\cite{Simonyan2014Very}. The coefficient $\lambda$ is the loss weight ($0.005$ by default).

\subsection{Warping Field Network} \label{subsec:warping}
This section introduces a warping field network, inspired by prior works~\cite{jiang2022neuman,weng2022humannerf,bai2023learning}, to correct the fitting noise in the real data.
The warping field network takes a \textit{trainable} per-frame latent variable (denoted as $\pmb{v}_i$, $i \in [N]$, $N$ is the number of training frames) and point position ($\pmb{q}$) as input, and output a new point position (\ie, the warped point position, denoted as $\pmb{q}'$) by the following formulation~\cite{bai2023learning},
\begin{equation}
\begin{aligned}
    \pmb{R}, \pmb{c}^{\mathrm{(rot)}}, \pmb{t} = \mathcal{G}(\pmb{q}, \pmb{v}_i), \\
    \pmb{q}' = \pmb{R}(\pmb{q} + \pmb{c}^{\mathrm{(rot)}}) - \pmb{c}^{\mathrm{(rot)}} + \pmb{t},
\end{aligned}
\end{equation}
where $\pmb{R}, \pmb{c}^{\mathrm{(rot)}}, \pmb{t} $ stands for rotation matrix, rotation center, translation, respectively; $\mathcal{G}$ is the warping network, which is designed as a residual MLP, similar to the LightAvatar backbone \textit{but much shallower} (we empirically find a deep warping field network is hard to converge). %

During training, we mix the pseudo and real images for the best performance. Notably, the warping field network is only used for the \textit{real} images. For pseudo images, since they are exactly aligned, they will not go through the warping field network. During testing, the warping field network is \textit{not} needed, either.

%% file: sec/4_experiments.tex
\section{Experimental Results}
\label{sec:experiments}

\textbf{Implementation Details.}
We use TensorFlow~\cite{abadi2016tensorflow} for the major experiments. Adam optimizer~\cite{kingma2014adam} is used with an exponential decay learning rate (LR) schedule (initial LR 5e-4, decayed by a multiplier $0.2$ every 500K iterations). During training, we train the model without SR network first with ray-based pseudo data (16,384 rays per batch) for around 500K iterations. This provides the initial weights. Next, we add the SR network to jointly optimize for another 500K iterations with initial LR 1e-4, where the training data is image-based (16 images per batch). When finetuning with the real data (mixed with pseudo data), the initial LR is set to even smaller (1e-5) to avoid over-optimization.

Our NeLF model has 10 residual MLP blocks (width 128 neurons). Each residual MLP block has 2 MLP layers. The SR model has 5 residual Conv blocks (width 56 filters, kernel size $3\times3$, padding 1) with two $\times2$ upsample layers (thus, the total upsample scale is $\times4$). The total FLOPs of the whole model is designed to be ultra low: \textbf{0.09M per pixel}, which is an order-of-magnitude smaller than existing counterparts (see Tab.~\ref{tab:speed_comparison}). Code is released at: \href{https://github.com/MingSun-Tse/LightAvatar-TensorFlow}{https://github.com/MingSun-Tse/LightAvatar-TensorFlow}.

\vspace{0.3em}
\noindent \textbf{Datasets}. We compare our approach to others on \textbf{14} monocular videos of different subjects, named \textit{Subject0} to \textit{Subject13} in this work. \textit{Subject0} to \textit{Subject12} are from prior works like NerFACE~\cite{gafni2021dynamic} and MonoAvatar~\cite{bai2023learning}. \textit{Subject13} is captured by this work, with shoulder -- we shall show our method can reliably model the shoulder on this subject. The backgrounds of these videos are removed and only heads (and shoulder for \textit{Subject13}) are retained. We resize the video to make sure the longer side is 512. Each video is split into two parts. The first part is used for training and the other part reserved for testing. We choose MonoAvatar~\cite{bai2023learning} as teacher to synthesize $20K$ pseudo frames~(Sec.~\ref{subsec:training_via_kd}).

\vspace{0.3em}
\noindent \textbf{Comparison Methods and Evaluation Metrics}. We compare with existing popular head avatar methods: FOMM~\cite{siarohin2019first}, TPSMM~\cite{zhao2022thin},  NHA~\cite{grassal2022neural}, IMAvatar~\cite{zheng2022avatar}, NerFACE~\cite{gafni2021dynamic}, and MonoAvatar~\cite{bai2023learning}.  Among them, MonoAvatar is the prior SOTA in terms of quality (thus chosen as our teacher). Besides, some very recent works also focus on fast avatars as we do, \eg, NeRFBlendShape~\cite{gao2022reconstructing}, PointAvatar~\cite{zheng2023pointavatar}, and INSTA~\cite{zielonka2023instant}. Our work is fundamentally different from them in that we build upon a different representation (neural light fields instead of NeRFs or point clouds). We shall also compare with them.

We evaluate different methods with standard metrics: LPIPS~\cite{zhang2018unreasonable} / SSIM~\cite{wang2004image} / PSNR. Of note, it is well-known that LPIPS capture the structural details more accurately than the pixel-wise PSNR and patch-based SSIM. Thus, when evaluating the quantitative results, it is advised to put more weight on LPIPS.

\subsection{Comparison with Other Approaches}
\vspace{0.3em}
\noindent \textbf{Quantitative Comparison}. The quantitative comparisons are presented in Tab.~\ref{tab:results_quantitative_flame} and Tab.~\ref{tab:quantitative_results_fast_avatars}. Our LightAvatar consistently achieves the \textit{best} average LPIPS/SSIM/PSNR in two tables, showing the encouraging potential of employing light fields to represent head avatars.

Notably, although we use MonoAvatar~\cite{bai2023learning} as teacher to synthesize pseudo data, our final results actually \textit{surpass} the teacher (see Tab.~\ref{tab:results_quantitative_flame}). This phenomenon agrees with the previous observations in the static neural light fields~\cite{wang2022r2l,cao2023real}. This is because the pseudo data (\ie, the supervision from the teacher) only provides the initial weights to our model for the subsequent finetuning on the real data. The performance of our model is \textit{not} bounded by the teacher.

\begin{table}[t]
\centering
\caption{LPIPS$\downarrow$/SSIM$\uparrow$/PSNR$\uparrow$ comparison with prior qualitatively top-performing methods. The \textbf{best} results are \textbf{in bold}, \ul{second best} \ul{underlined}.}
\vspace{-2mm}
\resizebox{0.995\linewidth}{!}{
\setlength{\tabcolsep}{0.3mm}
\begin{tabular}{l|c|c|c|c|c|c}
\multirow{2}{*}{Method} & \textit{Subject 0} & \textit{Subject 1} & \textit{Subject 2} & \textit{Subject 3} & \textit{Subject 4} & \textit{Average} \\
& LPIPS/SSIM/PSNR & LPIPS/SSIM/PSNR & LPIPS/SSIM/PSNR & LPIPS/SSIM/PSNR & LPIPS/SSIM/PSNR & LPIPS/SSIM/PSNR \\
\Xhline{3\arrayrulewidth}
TPSMM & 0.192/0.852/22.60 & 0.205/0.830/16.38 & 0.216/0.782/18.40 & 0.222/0.799/20.28 & 0.156/0.913/21.29 & 0.198/0.835/19.79 \\
FOMM & 0.171/0.841/22.93 & 0.179/0.827/16.02 & 0.202/0.777/18.98 & 0.186/0.798/22.28 & 0.122/0.915/23.94 & 0.172/0.832/20.83 \\
NHA & 0.165/0.836/20.20 & 0.166/0.840/15.48 & 0.178/0.809/17.99 & 0.153/0.798/21.31 & 0.091/0.926/23.78 & 0.151/0.842/19.75 \\
IMAvatar & 0.207/0.852/21.26 & 0.187/0.848/15.98 & 0.265/0.729/15.80 & 0.214/0.782/20.37 & 0.142/0.897/20.63 & 0.203/0.822/18.81 \\
NerFACE & 0.205/0.817/20.06 & 0.182/0.833/15.78 & 0.188/0.793/19.41 & 0.229/0.747/18.16 & 0.093/0.938/25.57 & 0.179/0.826/19.80 \\
MonoAvatar & 0.144/0.864/21.92 & 0.152/0.855/16.23 & 0.141/0.841/20.42 & 0.156/0.833/23.05 & 0.075/0.944/25.71 & \ul{0.134}/\ul{0.867}/\ul{21.47} \\
\rowcolor[gray]{0.92} Ours & 0.136/0.864/22.12 & 0.138/0.855/16.32 & 0.117/0.844/20.82 & 0.137/0.836/23.43 & 0.060/0.947/25.88 & \textbf{0.118}/\textbf{0.869}/\textbf{21.71} \\
\end{tabular}}
\vspace{-5mm}
\label{tab:results_quantitative_flame}
\end{table}

\begin{figure*}[t]
\centering
\resizebox{0.995\linewidth}{!}{
\setlength{\tabcolsep}{0mm}
\renewcommand{\arraystretch}{0.01}
\begin{tabular}{ccccccc}
\includegraphics[trim=1cm 1cm 1cm 1cm,clip,width=0.141\linewidth]{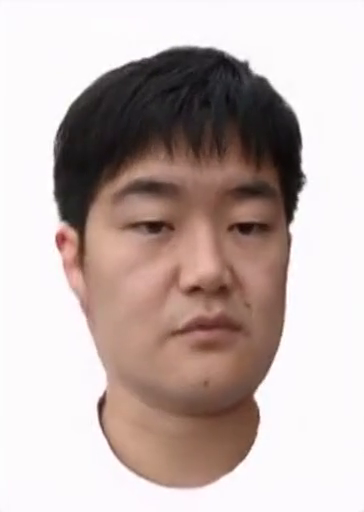} &
\includegraphics[trim=1cm 1cm 1cm 1cm,clip,width=0.141\linewidth]{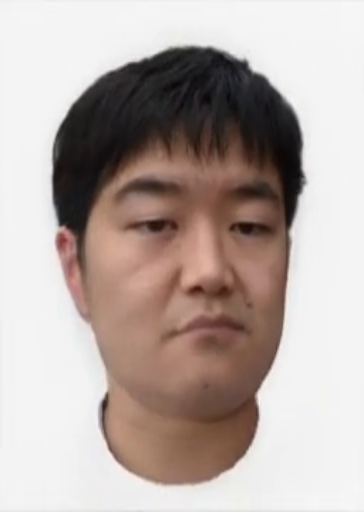} &
\includegraphics[trim=1cm 1cm 1cm 1cm,clip,width=0.141\linewidth]{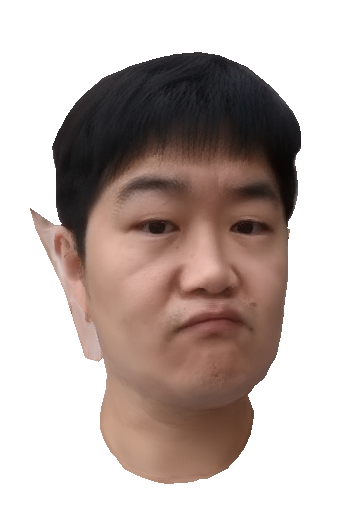} &
\includegraphics[trim=1cm 1cm 1cm 1cm,clip,width=0.141\linewidth]{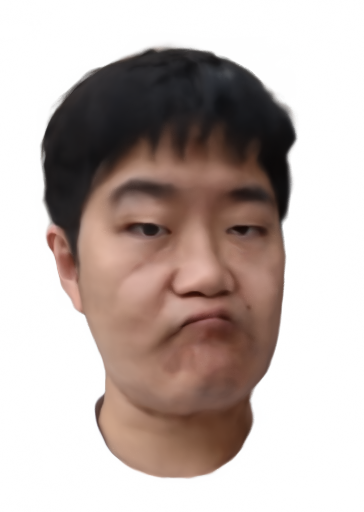} & 
\includegraphics[trim=1cm 1cm 1cm 1cm,clip,width=0.141\linewidth]{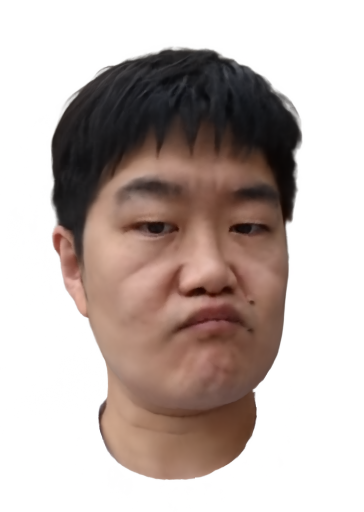} &
\includegraphics[trim=1cm 1cm 1cm 1cm,clip,width=0.141\linewidth]{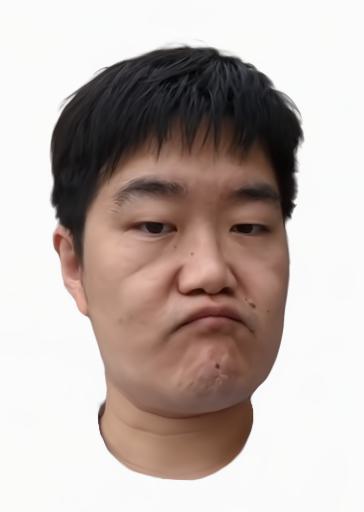} &
\includegraphics[trim=1cm 1cm 1cm 1cm,clip,width=0.141\linewidth]{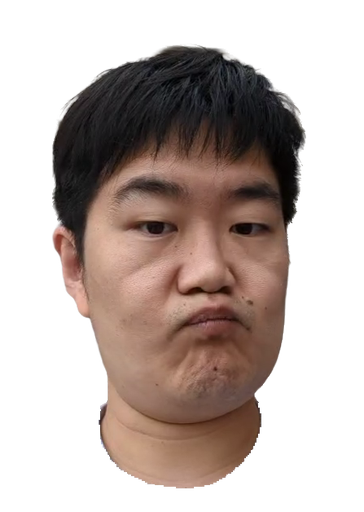} \\
\includegraphics[trim=1cm 1cm 1cm 1cm,clip,width=0.141\linewidth]{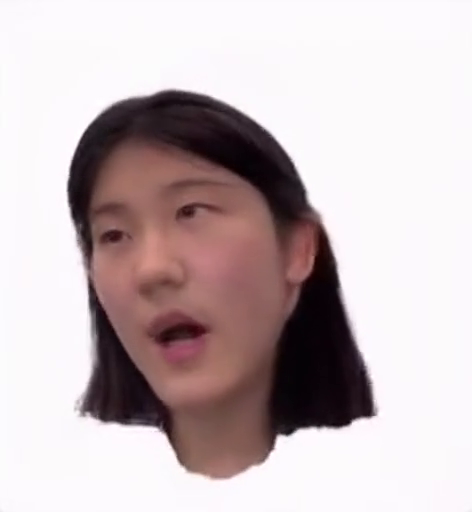} &
\includegraphics[trim=1cm 1cm 1cm 1cm,clip,width=0.141\linewidth]{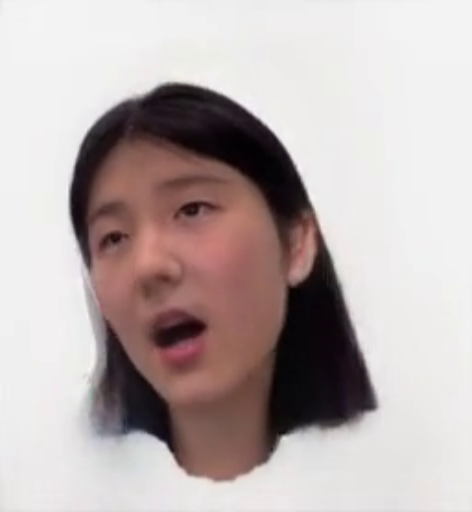} &
\includegraphics[trim=1cm 1cm 1cm 1cm,clip,width=0.141\linewidth]{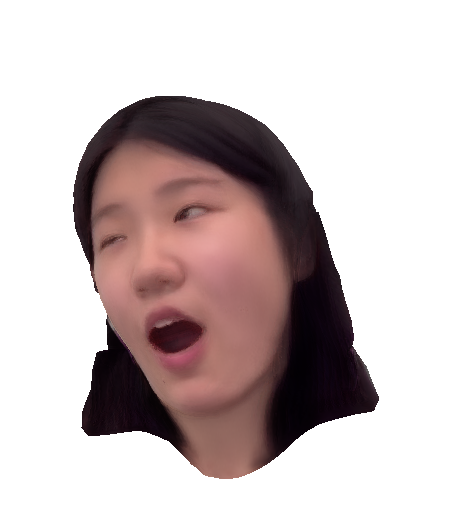} &
\includegraphics[trim=1cm 1cm 1cm 1cm,clip,width=0.141\linewidth]{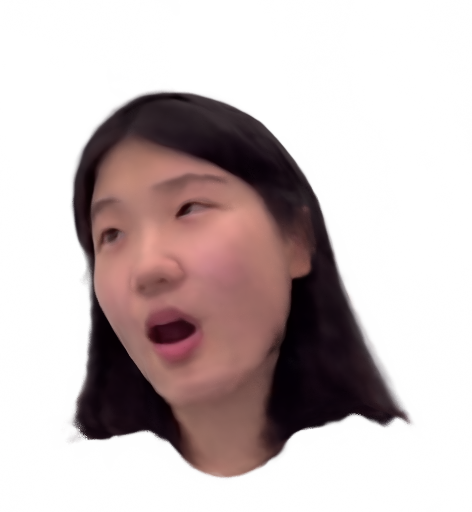} & 
\includegraphics[trim=1cm 1cm 1cm 1cm,clip,width=0.141\linewidth]{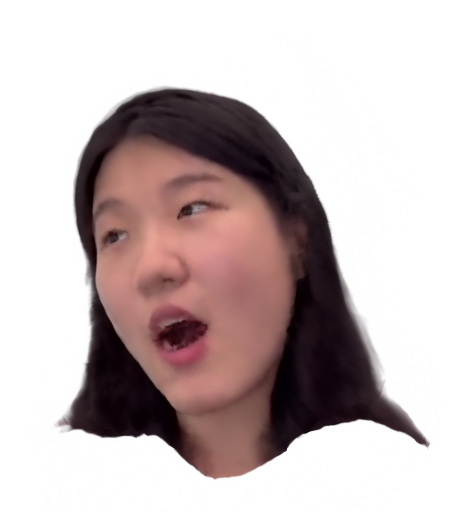} &
\includegraphics[trim=1cm 1cm 1cm 1cm,clip,width=0.141\linewidth]{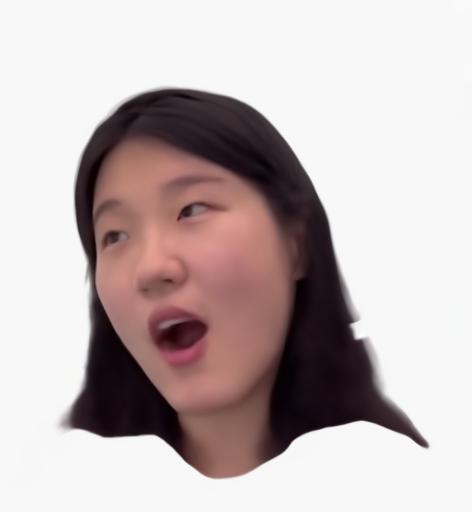} &
\includegraphics[trim=1cm 1cm 1cm 1cm,clip,width=0.141\linewidth]{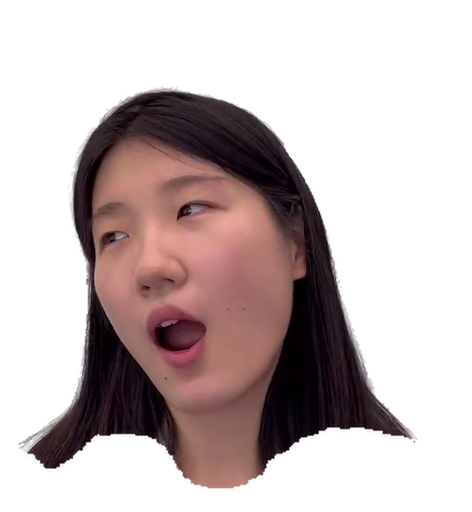} \\
\includegraphics[trim=1cm 1cm 1cm 1cm,clip,width=0.141\linewidth]{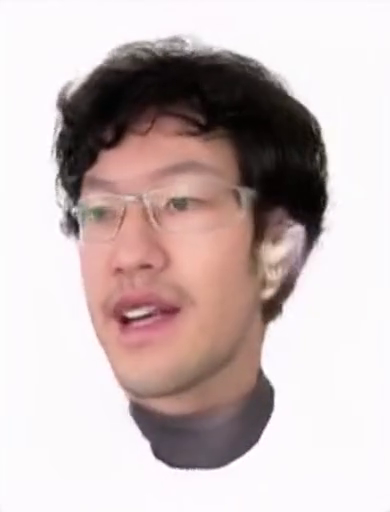} &
\includegraphics[trim=1cm 1cm 1cm 1cm,clip,width=0.141\linewidth]{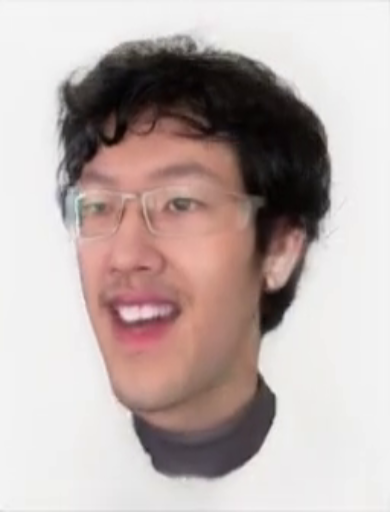} &
\includegraphics[trim=1cm 1cm 1cm 1cm,clip,width=0.141\linewidth]{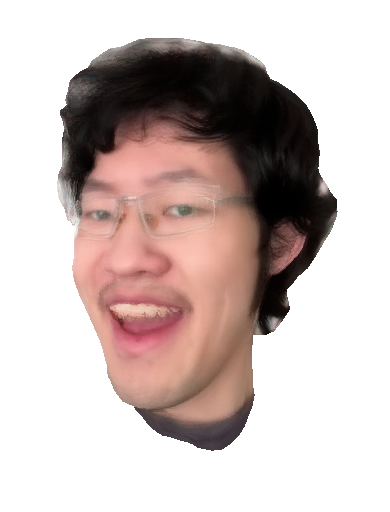} &
\includegraphics[trim=1cm 1cm 1cm 1cm,clip,width=0.141\linewidth]{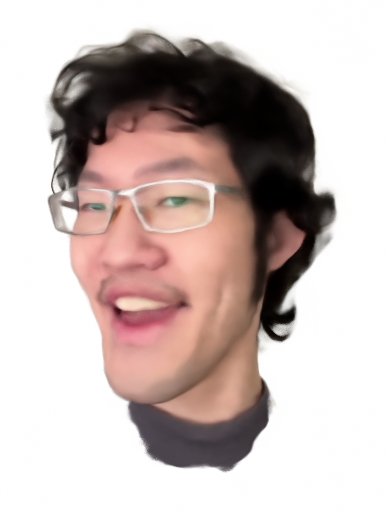} & 
\includegraphics[trim=1cm 1cm 1cm 1cm,clip,trim=1cm 1cm 1cm 1cm,clip,width=0.141\linewidth]{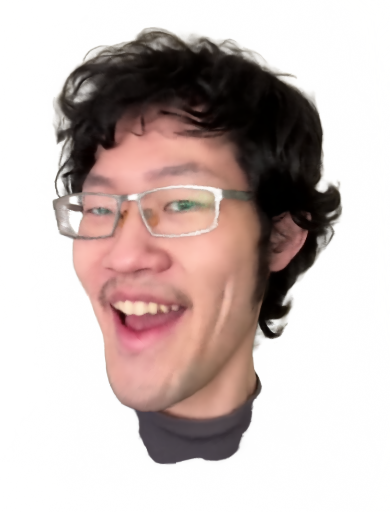} &
\includegraphics[trim=1cm 1cm 1cm 1cm,clip,width=0.141\linewidth]{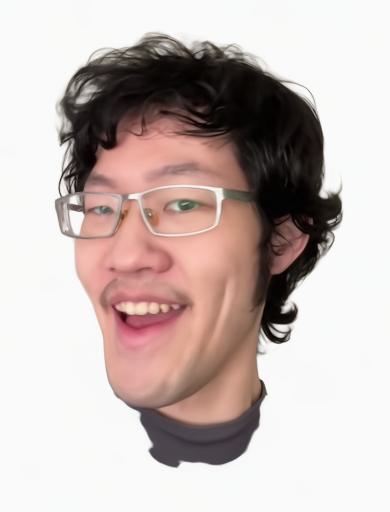} &
\includegraphics[trim=1cm 1cm 1cm 1cm,clip,width=0.141\linewidth]{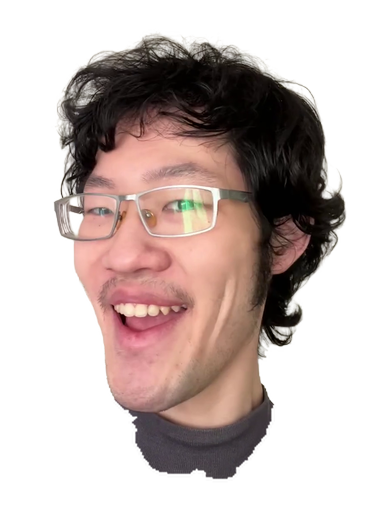} \\
\includegraphics[trim=1cm 0.5cm 1cm 1cm,clip,width=0.141\linewidth]{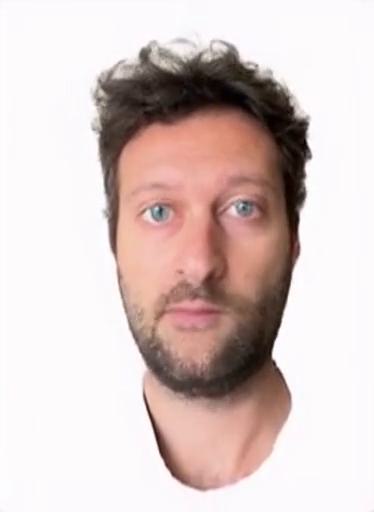} &
\includegraphics[trim=1cm 0.5cm 1cm 1cm,clip,width=0.141\linewidth]{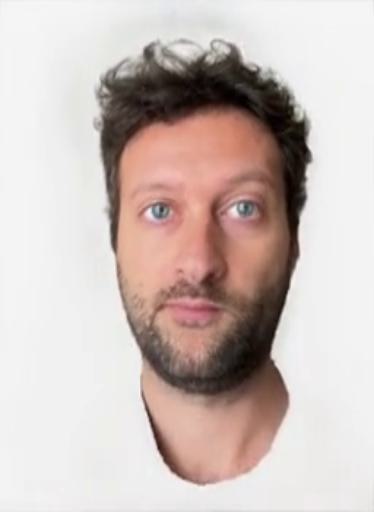} &
\includegraphics[trim=1cm 0.5cm 1cm 1cm,clip,width=0.141\linewidth]{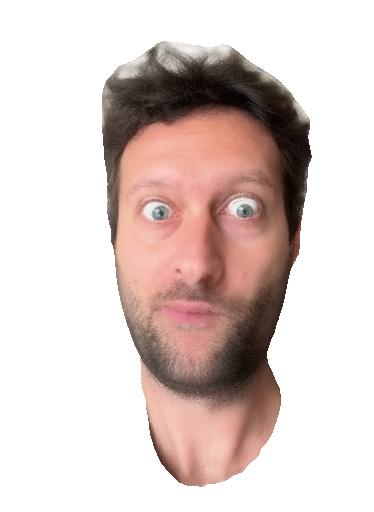} &
\includegraphics[trim=1cm 0.5cm 1cm 1cm,clip,width=0.141\linewidth]{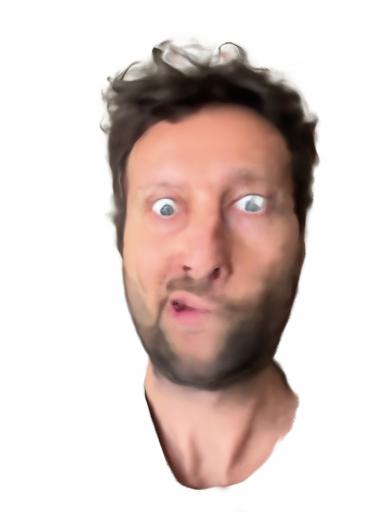} & 
\includegraphics[trim=1cm 0.5cm 1cm 1cm,clip,width=0.141\linewidth]{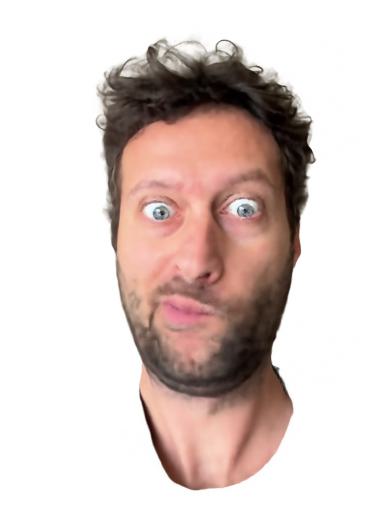} &
\includegraphics[trim=1cm 0.5cm 1cm 1cm,clip,width=0.141\linewidth]{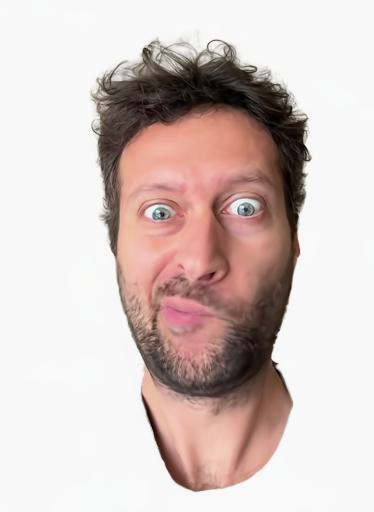} &
\includegraphics[trim=1cm 0.5cm 1cm 1cm,clip,width=0.141\linewidth]{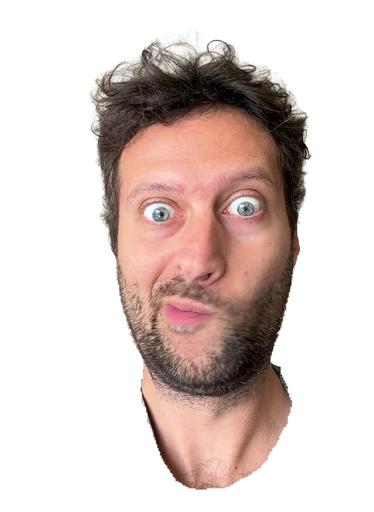} \\
\includegraphics[trim=2cm 0cm 2cm 2cm,clip,width=0.141\linewidth]{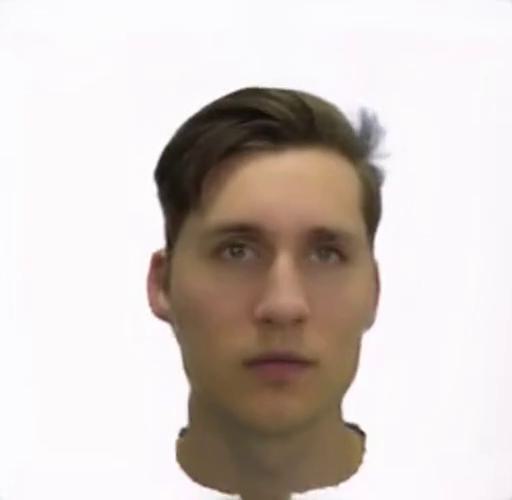} &
\includegraphics[trim=2cm 0cm 2cm 2cm,clip,width=0.141\linewidth]{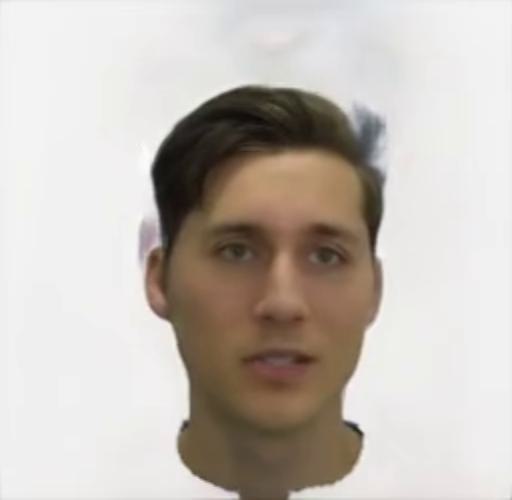} &
\includegraphics[trim=2cm 0cm 2cm 2cm,clip,width=0.141\linewidth]{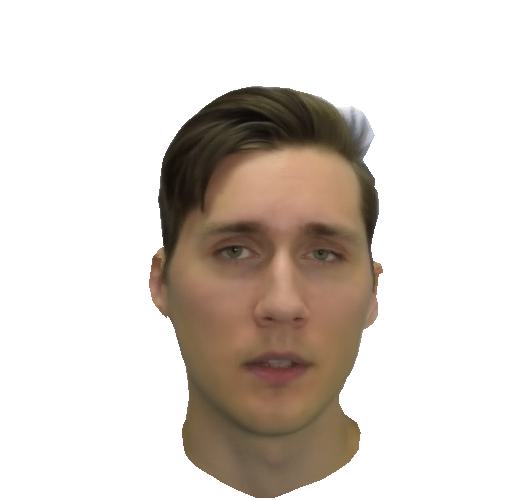} &
\includegraphics[trim=2cm 0cm 2cm 2cm,clip,width=0.141\linewidth]{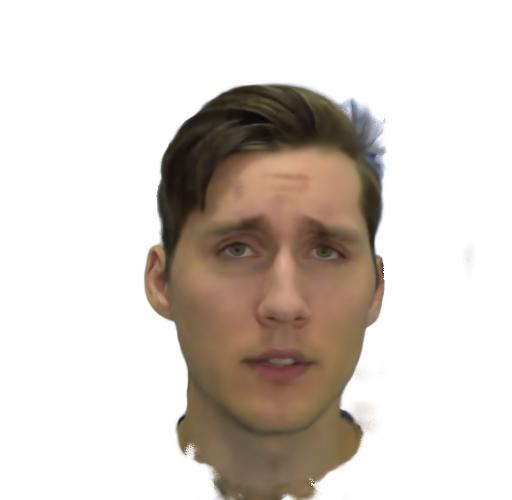} & 
\includegraphics[trim=2cm 0cm 2cm 2cm,clip,width=0.141\linewidth]{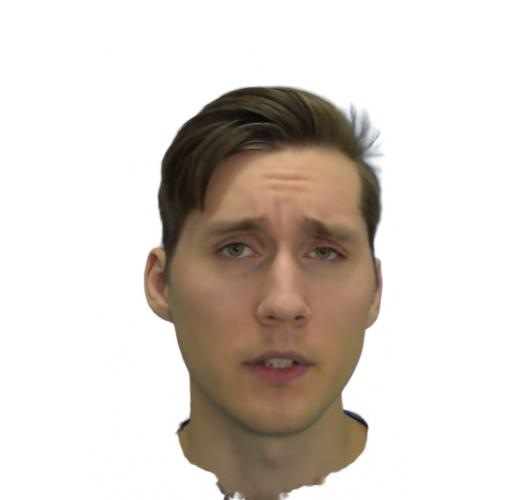} &
\includegraphics[trim=2cm 0cm 2cm 2cm,clip,width=0.141\linewidth]{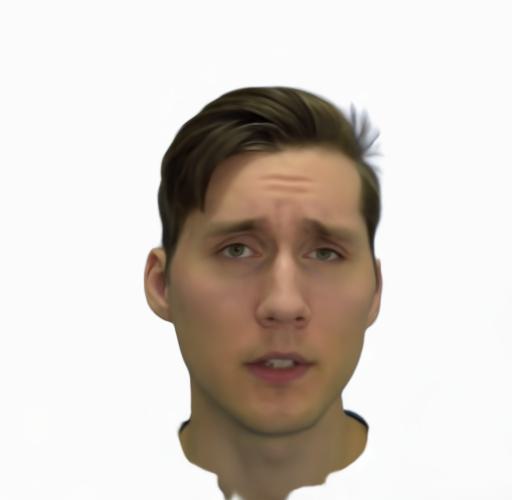} &
\includegraphics[trim=2cm 0cm 2cm 2cm,clip,width=0.141\linewidth]{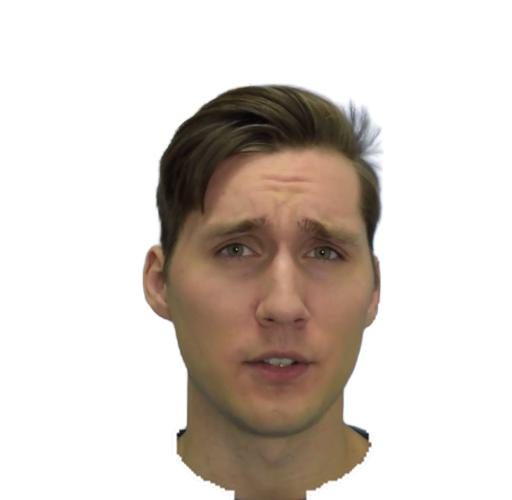} \\
\vspace{1mm} \\
\small FOMM~\cite{siarohin2019first} & \small TPSMM~\cite{zhao2022thin} & \small NHA~\cite{grassal2022neural} & \small NerFACE~\cite{gafni2021dynamic} & \small MonoAvatar~\cite{bai2023learning} & \small Ours & \small GT
\end{tabular}}
\caption{Visual comparison on the test set with prior top-performing monocular head avatars. From top to down, the subject is \textit{Subject0} to \textit{Subject4} in order (see another 3 in supplementary material). Our LightAvatar method faithfully predicts the facial expressions and presents \textit{sharper} high-frequency details than other approaches.}
\label{fig:results_qualitative_flame}
\vspace{-5mm}
\end{figure*}

\begin{table}[t]
\centering
\caption{LPIPS$\downarrow$/SSIM$\uparrow$/PSNR$\uparrow$ comparison with recent \textit{fast avatar} methods. NBS: NeRFBlendShape~\cite{gao2022reconstructing}, PA: PointAvatar~\cite{zheng2023pointavatar}, INSTA~\cite{zielonka2023instant}.}
\vspace{-2mm}
\resizebox{0.995\linewidth}{!}{
\setlength{\tabcolsep}{0.3mm}
\begin{tabular}{l|c|c|c|c|c|c}
\multirow{2}{*}{Method} & \textit{Subject 8} & \textit{Subject 9} & \textit{Subject 10} & \textit{Subject 11} & \textit{Subject 12} & \textit{Average} \\
& LPIPS/SSIM/PSNR & LPIPS/SSIM/PSNR & LPIPS/SSIM/PSNR & LPIPS/SSIM/PSNR & LPIPS/SSIM/PSNR & LPIPS/SSIM/PSNR \\
\Xhline{3\arrayrulewidth}
NBS & 0.093/0.882/20.50	& 0.104/0.933/25.99	& 0.108/0.900/25.66 & 0.081/0.924/24.78 &	0.049/0.960/23.78 &  \ul{0.087}/\ul{0.920}/\ul{24.14} \\
PA & 0.109/0.839/19.51 & 0.119/0.913/24.20 & 0.132/0.853/22.51 & 0.104/0.903/23.11 &	0.068/0.939/22.98 & 0.106/0.889/22.46 \\
INSTA & 0.124/0.857/20.65 &	0.142/0.910/23.3 &	0.173/0.833/20.39 & 0.102/0.913/24.28 & 0.058/0.951/23.93 & 0.120/0.893/22.51 \\
\rowcolor[gray]{0.92} Ours & 0.067/0.888/21.50 & 0.098/0.936/25.70 & 0.093/0.896/24.97 & 0.072/0.929/25.33 & 0.045/0.960/23.94 & \textbf{0.075/0.922/24.29} \\
\end{tabular}
}
\label{tab:quantitative_results_fast_avatars}
\vspace{-4mm}
\end{table}

\begin{figure*}[!h]
\centering
\resizebox{0.9\linewidth}{!}{
\setlength{\tabcolsep}{3mm}
\renewcommand{\arraystretch}{0.01}
\begin{tabular}{ccccccc}
\includegraphics[trim=1cm 0.5cm 1cm 2cm,clip,width=0.191\linewidth]{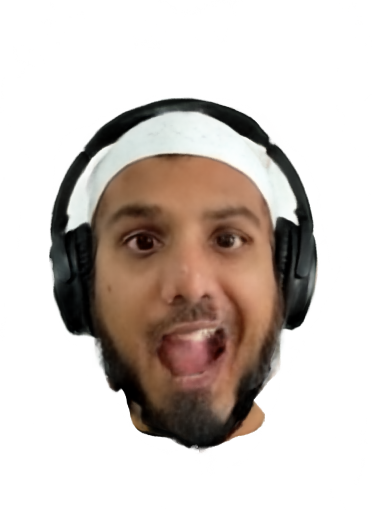} &
\includegraphics[trim=1cm 0.5cm 1cm 2cm,clip,width=0.191\linewidth]{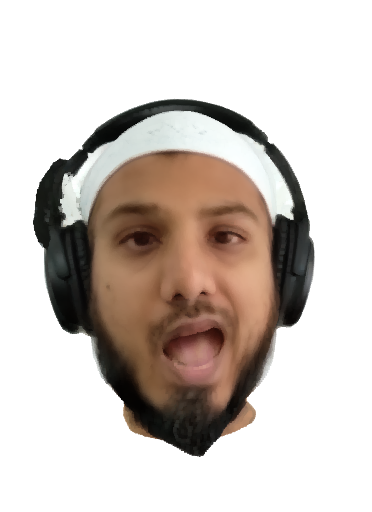} & 
\includegraphics[trim=1cm 0.5cm 1cm 2cm,clip,width=0.191\linewidth]{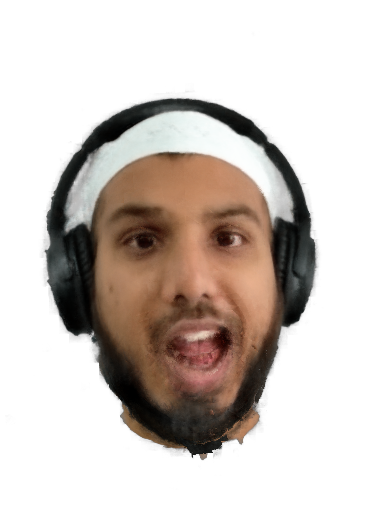} &
\includegraphics[trim=1cm 0.5cm 1cm 2cm,clip,width=0.191\linewidth]{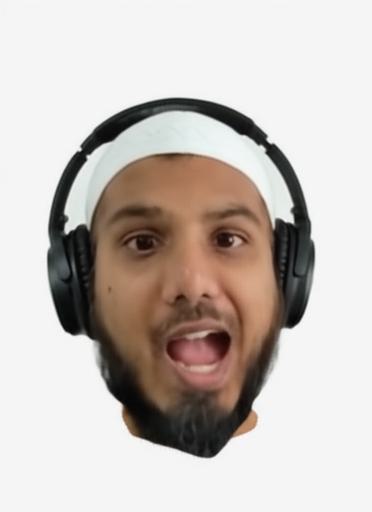} &
\includegraphics[trim=1cm 0.5cm 1cm 2cm,clip,width=0.191\linewidth]{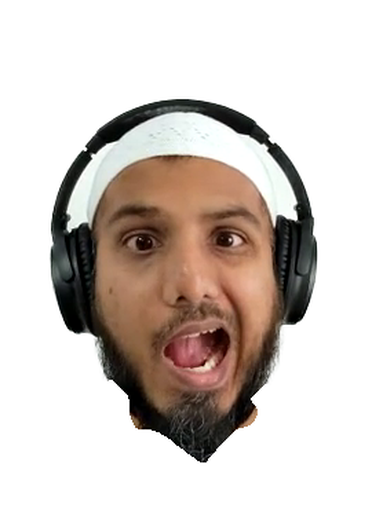} \\
\includegraphics[trim=2cm 2cm 2cm 2cm,clip,width=0.191\linewidth]{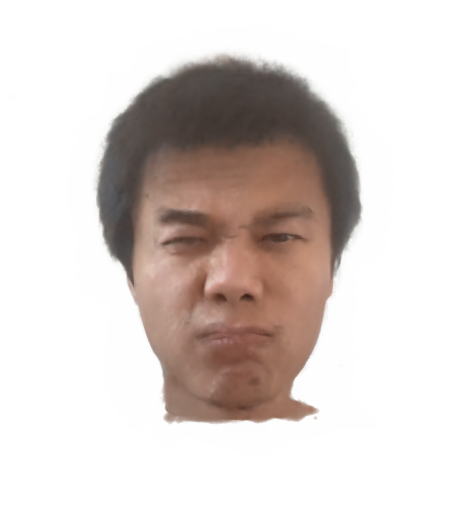} &
\includegraphics[trim=2cm 2cm 2cm 2cm,clip,width=0.191\linewidth]{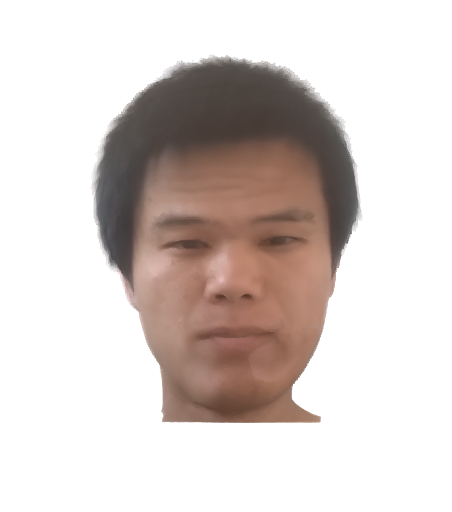} & 
\includegraphics[trim=2cm 2cm 2cm 2cm,clip,width=0.191\linewidth]{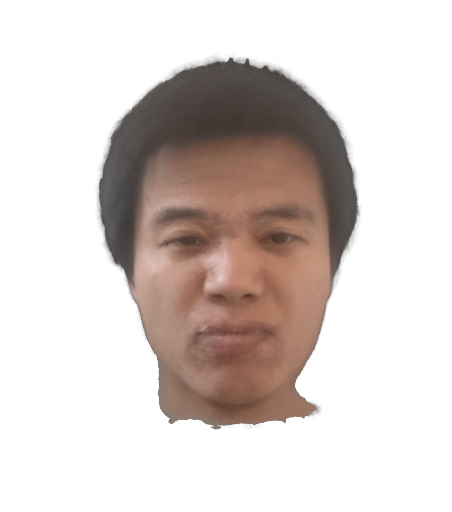} &
\includegraphics[trim=2cm 2cm 2cm 2cm,clip,width=0.191\linewidth]{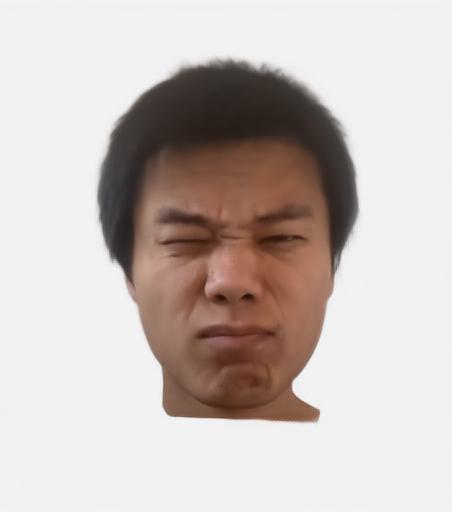} &
\includegraphics[trim=2cm 2cm 2cm 2cm,clip,width=0.191\linewidth]{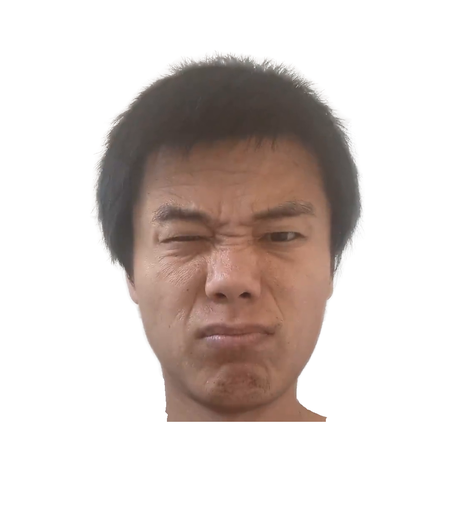} \\
\includegraphics[trim=2cm 0cm 1.5cm 1cm,clip,width=0.191\linewidth]{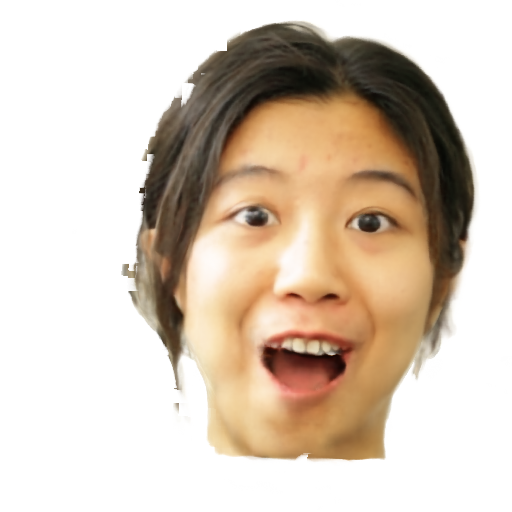} &
\includegraphics[trim=2cm 0cm 1.5cm 1cm,clip,width=0.191\linewidth]{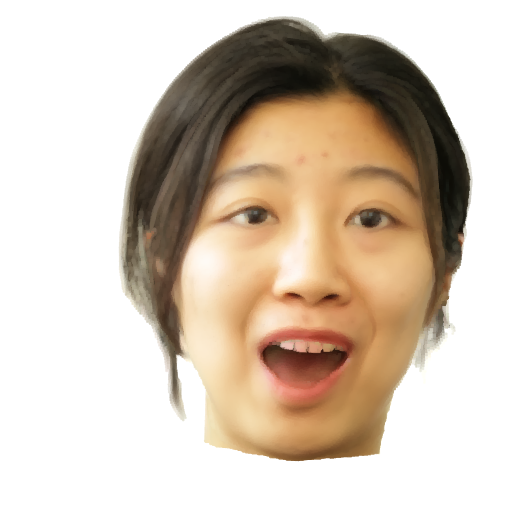} & 
\includegraphics[trim=2cm 0cm 1.5cm 1cm,clip,width=0.191\linewidth]{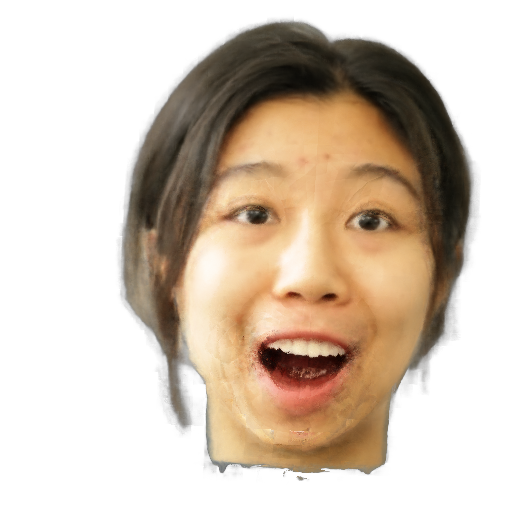} &
\includegraphics[trim=2cm 0cm 1.5cm 1cm,clip,width=0.191\linewidth]{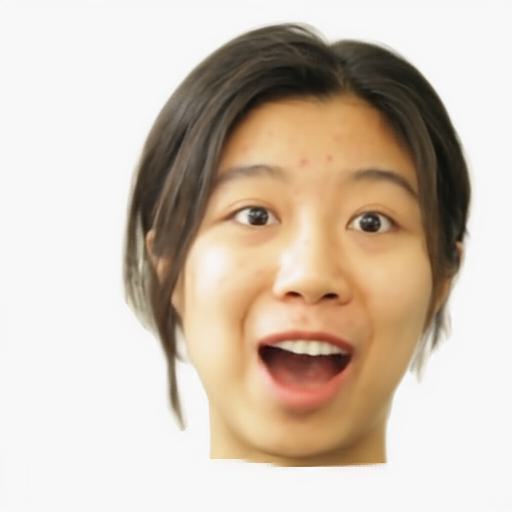} &
\includegraphics[trim=2cm 0cm 1.5cm 1cm,clip,width=0.191\linewidth]{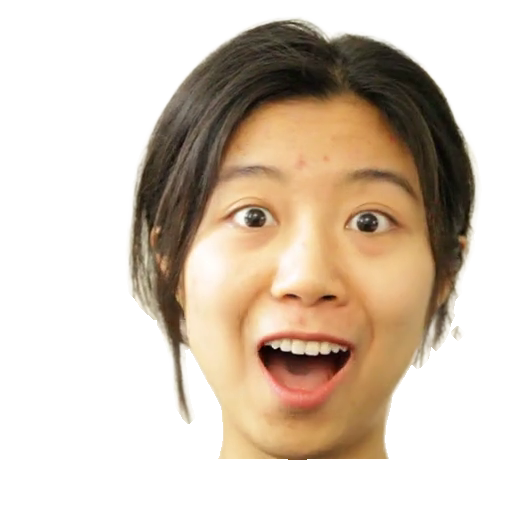} \\
\includegraphics[trim=0cm 3cm 1cm 2cm,clip,width=0.191\linewidth]{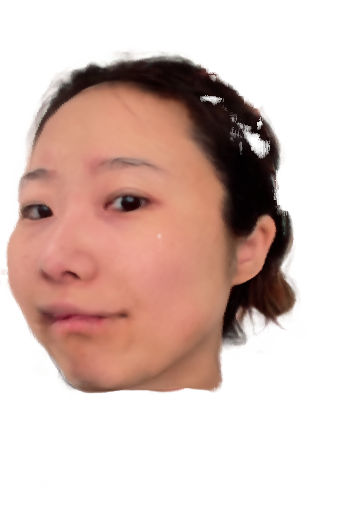} &
\includegraphics[trim=0cm 3cm 1cm 2cm,clip,width=0.191\linewidth]{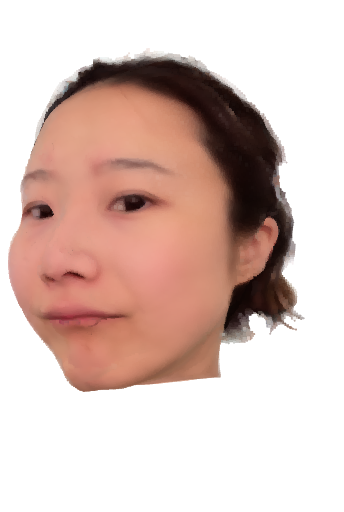} & 
\includegraphics[trim=0cm 3cm 1cm 2cm,clip,width=0.191\linewidth]{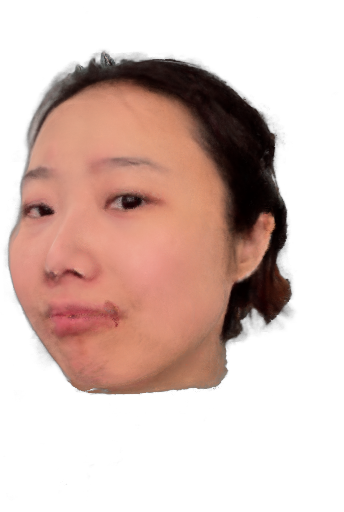} &
\includegraphics[trim=0cm 3cm 1cm 2cm,clip,width=0.191\linewidth]{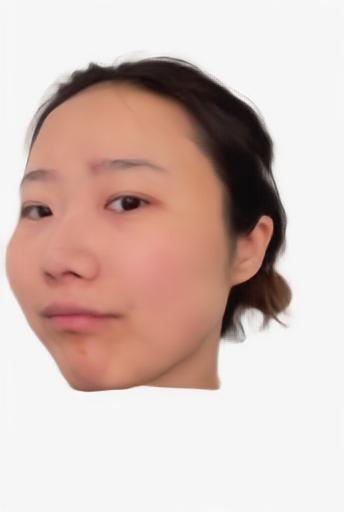} &
\includegraphics[trim=0cm 3cm 1cm 2cm,clip,width=0.191\linewidth]{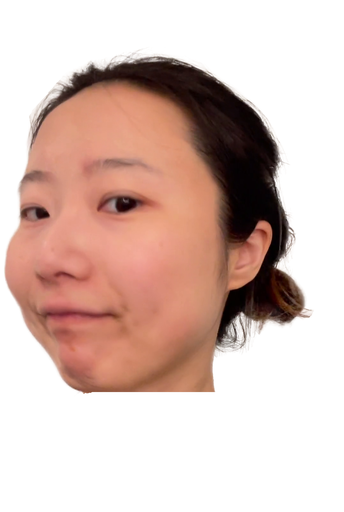} \\
\includegraphics[trim=3cm 4cm 2cm 0cm,clip,width=0.191\linewidth]{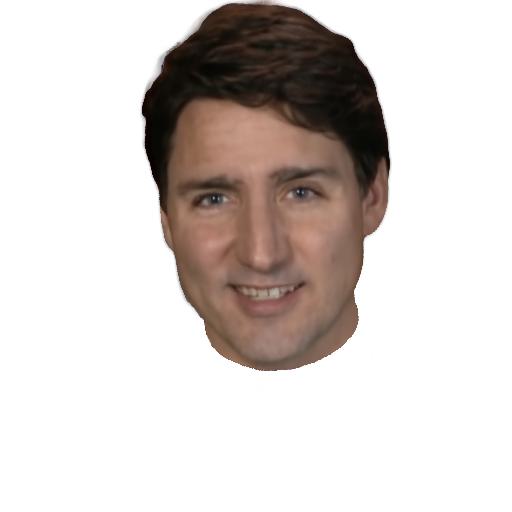} &
\includegraphics[trim=3cm 4cm 2cm 0cm,clip,width=0.191\linewidth]{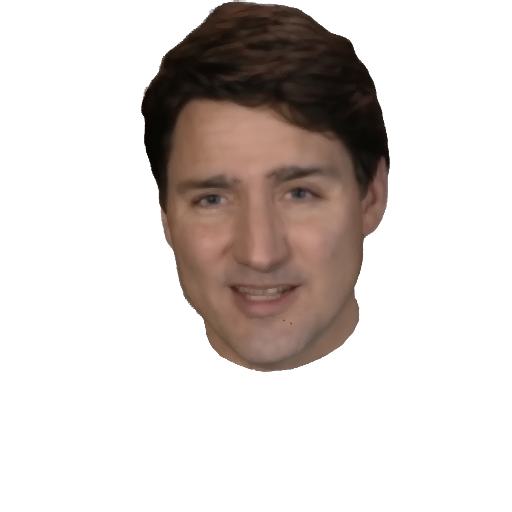} & 
\includegraphics[trim=3cm 4cm 2cm 0cm,clip,width=0.191\linewidth]{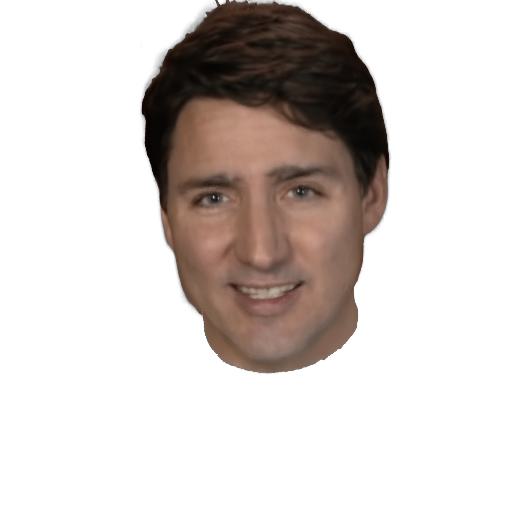} &
\includegraphics[trim=3cm 4cm 2cm 0cm,clip,width=0.191\linewidth]{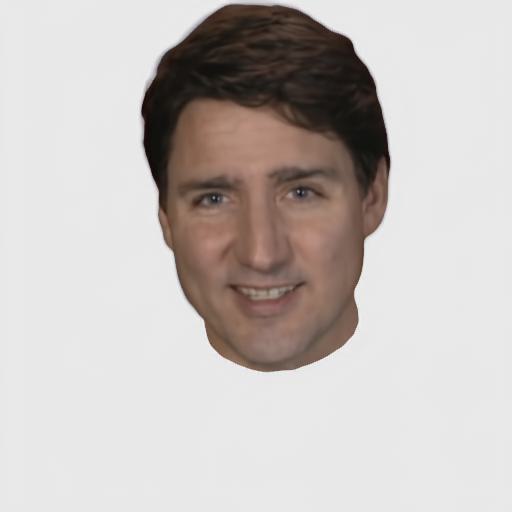} &
\includegraphics[trim=3cm 4cm 2cm 0cm,clip,width=0.191\linewidth]{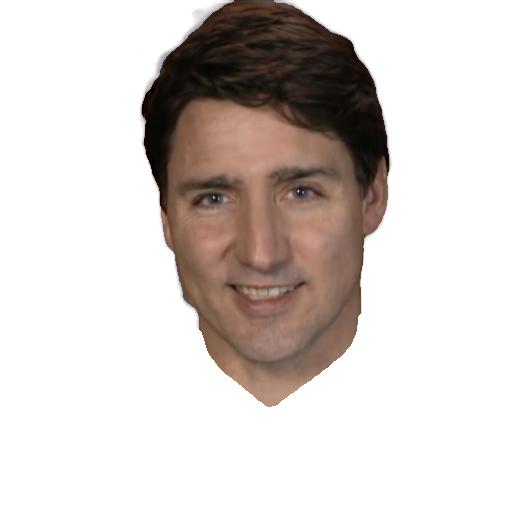} \\
\vspace{0.3em}
NBS & PA & INSTA & Ours & GT \\
\end{tabular}}
\caption{Visual comparison with recent fast avatars. NBS: NeRFBlendShape~\cite{gao2022reconstructing}, PA: PointAvatar~\cite{zheng2023pointavatar}, INSTA~\cite{zielonka2023instant}. Top to down: \textit{Subject 8} to \textit{Subject 12}.}
\label{fig:results_qualitative_fast_avatars}
\vspace{-5mm}
\end{figure*}

\vspace{0.3em}
\noindent \textbf{Qualitative Comparison}. The qualitative results are shown in Fig.~\ref{fig:results_qualitative_flame} and Fig.~\ref{fig:results_qualitative_fast_avatars}. (1) As seen, although our method does not rely on explicit mesh, it faithfully predicts the nuanced facial expressions like those that explicitly use mesh (such as MonoAvatar~\cite{bai2023learning}). LightAvatar is slightly better than MonoAvatar in general, as it produces less grain noise (\eg, zoom in and compare the eyes and mouth region of \textit{Subject 1}). (2) Compared to the fast avatars, our LightAvatar generates obviously better quality. According to the results, PointAvatar~\cite{zheng2023pointavatar} often distorts the facial expressions. INSTA~\cite{zielonka2023instant} produces quite many artifacts around the mouth area. NeRFBlendShape~\cite{gao2022reconstructing} sometimes produce incomplete structures (like the hair in \textit{Subject 10} and \textit{Subject 11}).

\vspace{0.3em}
\noindent \textbf{Inference and Training Efficiency}. \textbf{(1)} Tab.~\ref{tab:speed_comparison} shows the model complexity (FLOPs and run-time speed) comparison. Our method is \textit{significantly faster} than the teacher MonoAvatar~\cite{bai2023learning} and other fast avatars methods, thanks to our dedicated light field design mixed with 2D SR ConvNet. Notably, even INSTA~\cite{zielonka2023instant} employs customized C++/CUDA implementation, which is usually faster than pure TensorFlow or PyTorch implementation, our method is still significantly faster than INSTA, owing to the ultra-low FLOPs.

\textbf{(2)} This work does not focus on improving the training speed. It takes around 20 hrs with 4 V100 (16GB) GPUs to train our model of one subject with TensorFlow, which is only $1/5$ cost of previous NeLF papers on static scenes~\cite{wang2022r2l}, thanks to our ultra lightweight model design. For reference, PointAvatar reports $\sim$6 hrs with 1 A100 (80G), comparable to our cost considering A100 is around $2\sim3\times$ faster than V100. INSTA markedly reports $\sim$10 mins with 1 RTX 3090 GPU, yet it comes with an inferior quality and slower inference than ours. 

\begin{table}[t]
\centering
\caption{FLOPs \ul{per pixel} (estimated for rendering a $512\times 512$ image) and run-time speed (fps) comparison on an NVIDIA GeForce RTX 3090 GPU (24GB). Compared to the teacher, we achieve speedup by \textit{two orders of magnitude}. Note, our method is not only much faster, but also superior in terms of image quality (see Tab.~\ref{tab:quantitative_results_fast_avatars}, Fig.~\ref{fig:results_qualitative_fast_avatars}).}
\vspace{-2mm}
\resizebox{0.9\linewidth}{!}{
\setlength{\tabcolsep}{3mm}
\begin{tabular}{lccc}
\multirow{1}{*}{Method} & \multirow{1}{*}{FLOPs (M)} & Implementation & Speed (fps) \\
\Xhline{3\arrayrulewidth}
MonoAvatar~\cite{bai2023learning} (teacher) & 9.10 & TensorFlow & 0.5 \\
PointAvatar~\cite{zheng2023pointavatar} & 3.56 & PyTorch & 5.0 \\
INSTA~\cite{zielonka2023instant} & 1.01 & C++/CUDA & \ul{46.2} \\
NeRFBlendShape~\cite{gao2022reconstructing} & \ul{0.85} & PyTorch & 11.2 \\
\rowcolor[gray]{0.92} Ours & \textbf{0.09} & PyTorch & \textbf{174.1} \\
\end{tabular}}
\label{tab:speed_comparison}
\vspace{-1mm}
\end{table}

\begin{figure}[t]
\centering
\resizebox{0.995\linewidth}{!}{
\setlength{\tabcolsep}{0.5mm} %
\begin{tabular}{cccccc}
\includegraphics[width=0.16\linewidth]{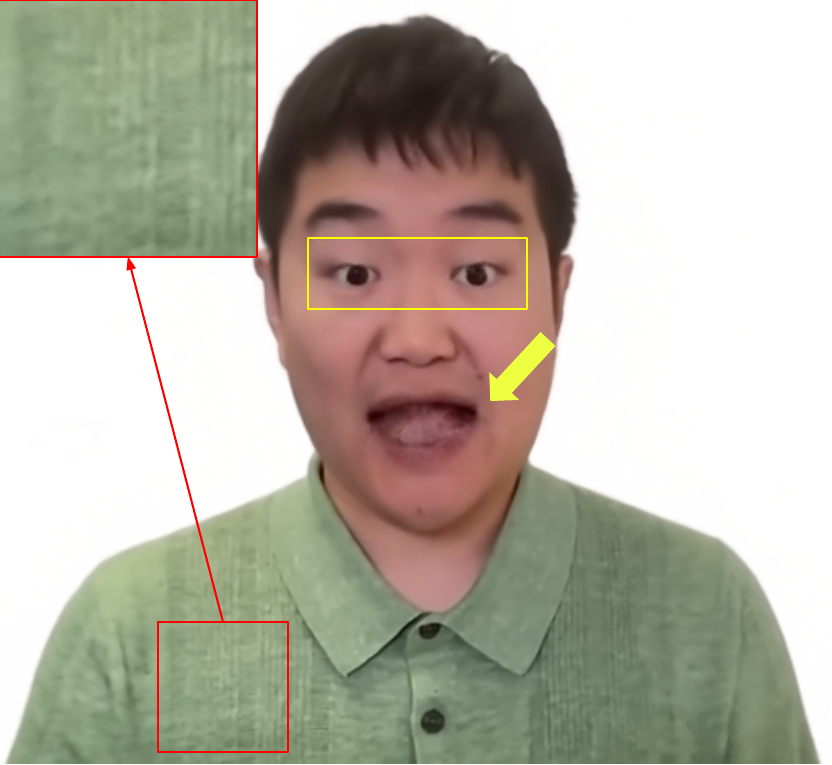} &
\includegraphics[width=0.16\linewidth]{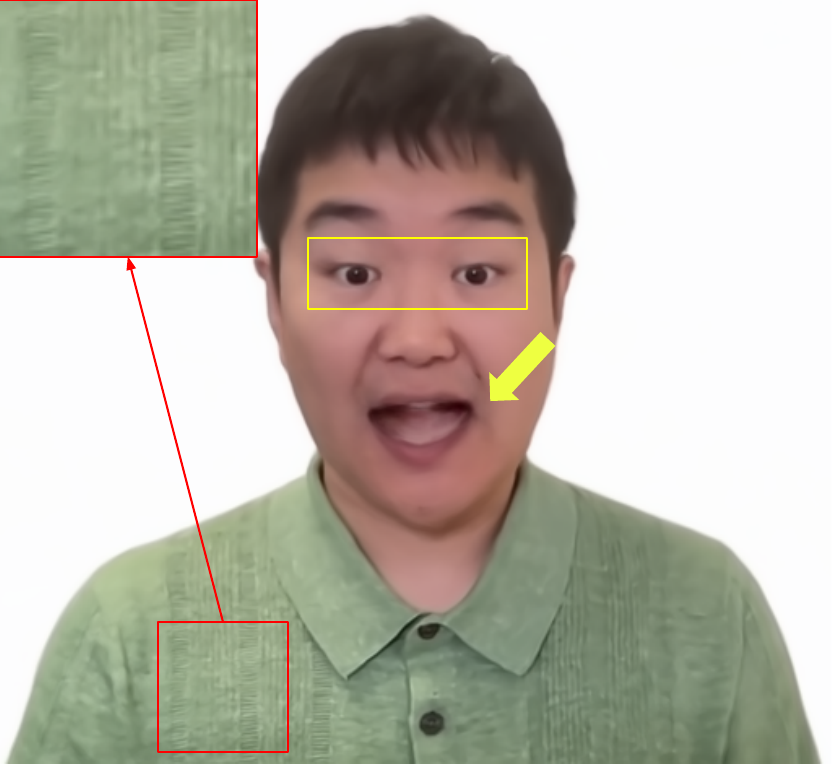} &
\includegraphics[width=0.16\linewidth]{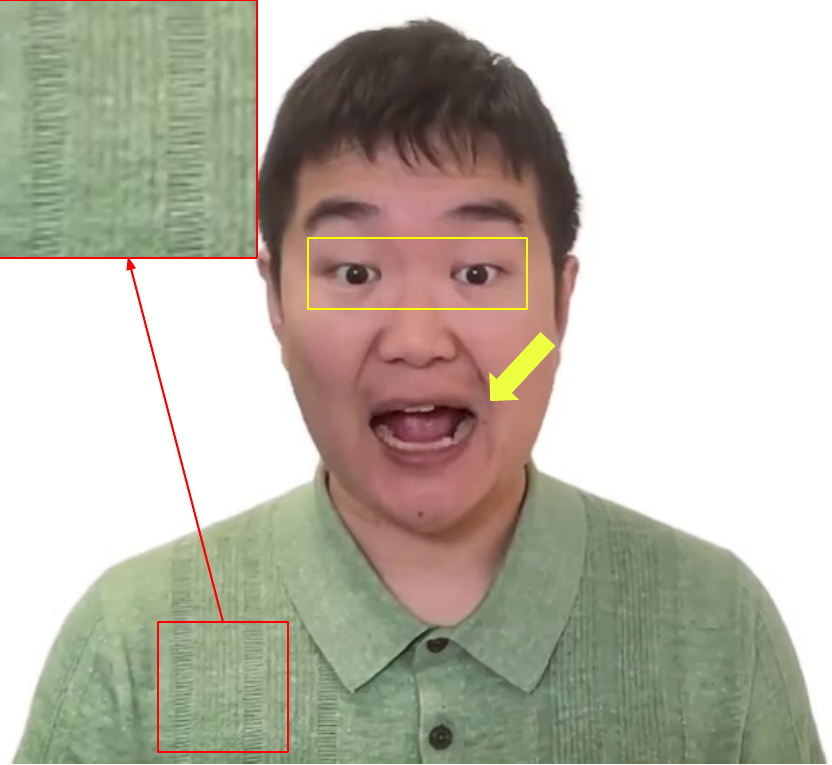} &
\includegraphics[width=0.16\linewidth]{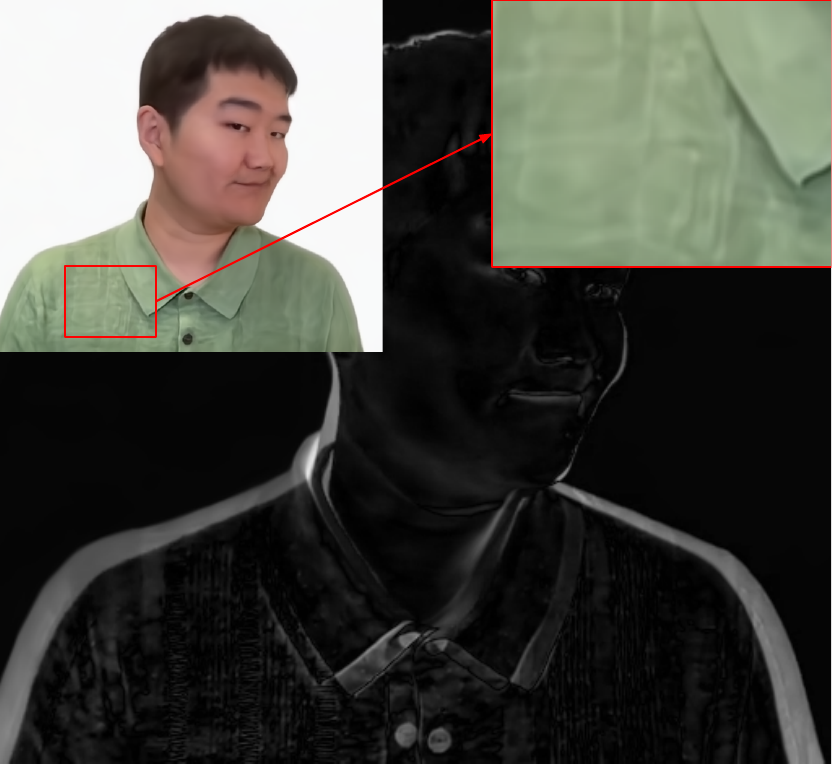} &
\includegraphics[width=0.16\linewidth]{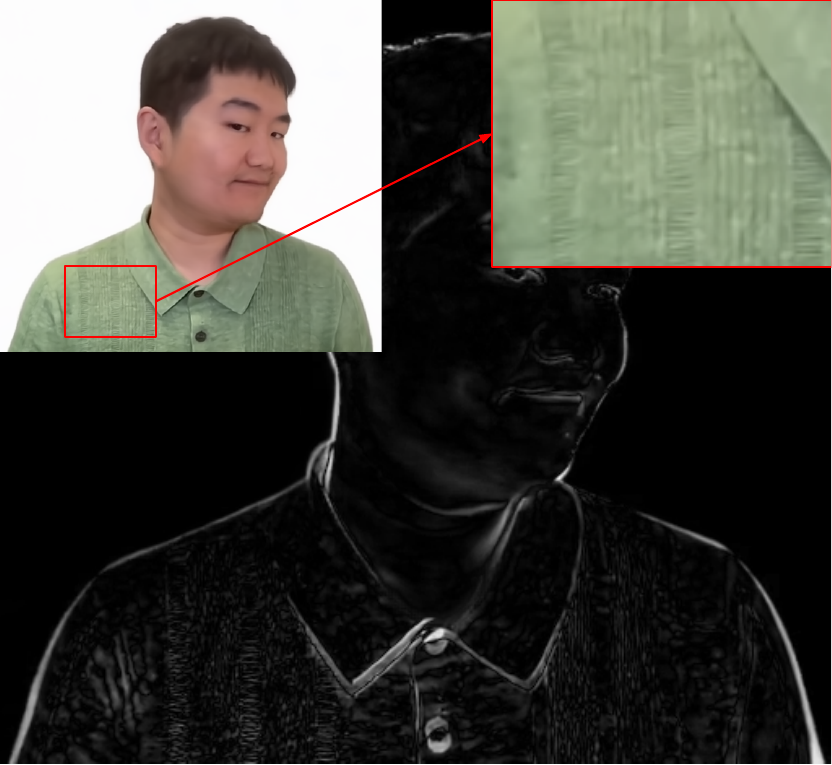} & 
\includegraphics[width=0.16\linewidth]{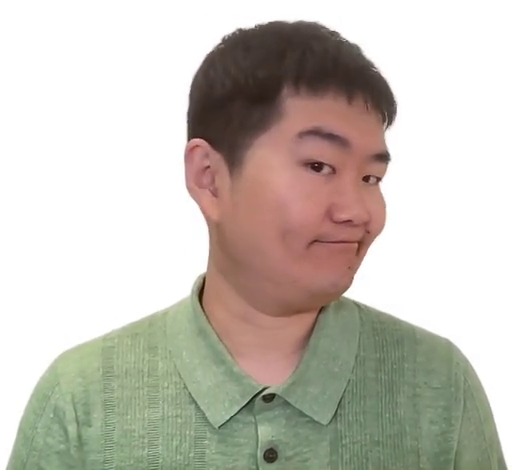} \\
\small (a.1) MonoAvatar & \small (a.2) Ours  & \small (a.3) GT & (b.1) Joint & (b.2) Separate & (b.3) GT  \\
\end{tabular}}
\vspace{-2mm}
\caption{(a) Results of our method on \textit{Subject13} with the shoulder. For reference, the average LPIPS/SSIM/PSNR of Monoavatar on test set: 0.118/0.846/26.11; ours: 0.107/0.849/26.29. (b) Comparison between \textit{joint modeling} and \textit{separate modeling} (Sec.~\ref{subsec:lightavatar}) when learning the shoulder in our method.}
\label{fig:results_with_shoulder}
\vspace{-4mm}
\end{figure}

\begin{figure}[t]
\centering
\setlength{\tabcolsep}{1mm} %
\begin{tabular}{cccccc}
\includegraphics[width=0.25\linewidth]{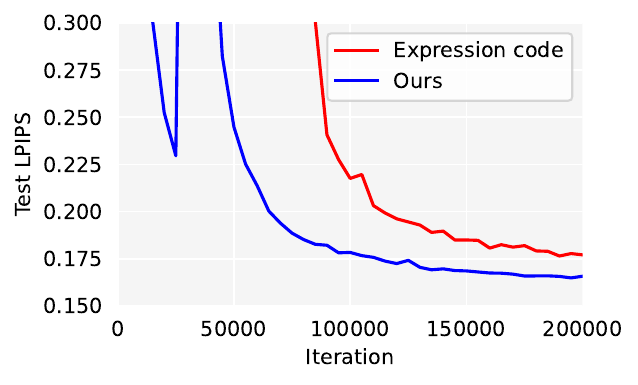} &
\includegraphics[width=0.16\linewidth]{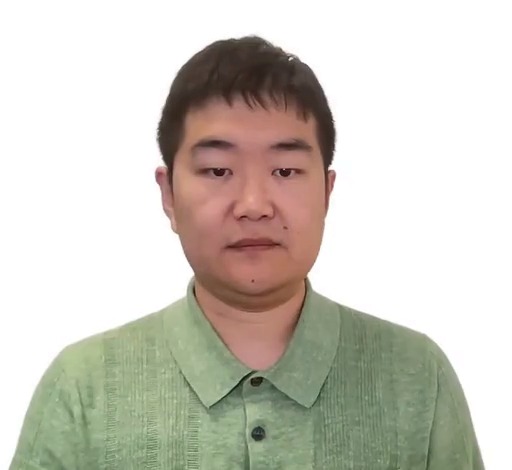} &
\includegraphics[width=0.16\linewidth]{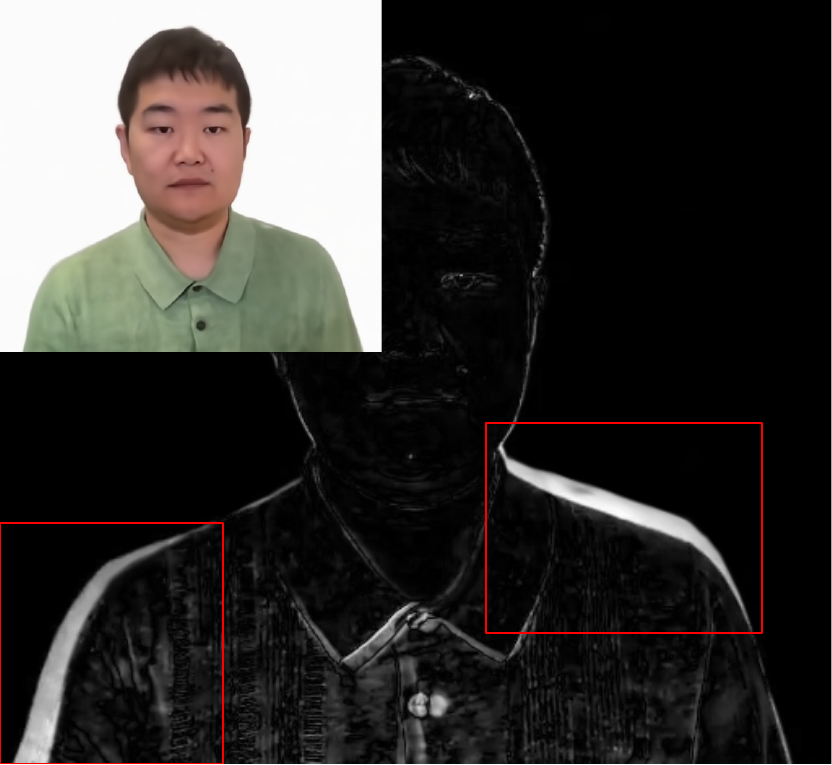} &
\includegraphics[width=0.16\linewidth]{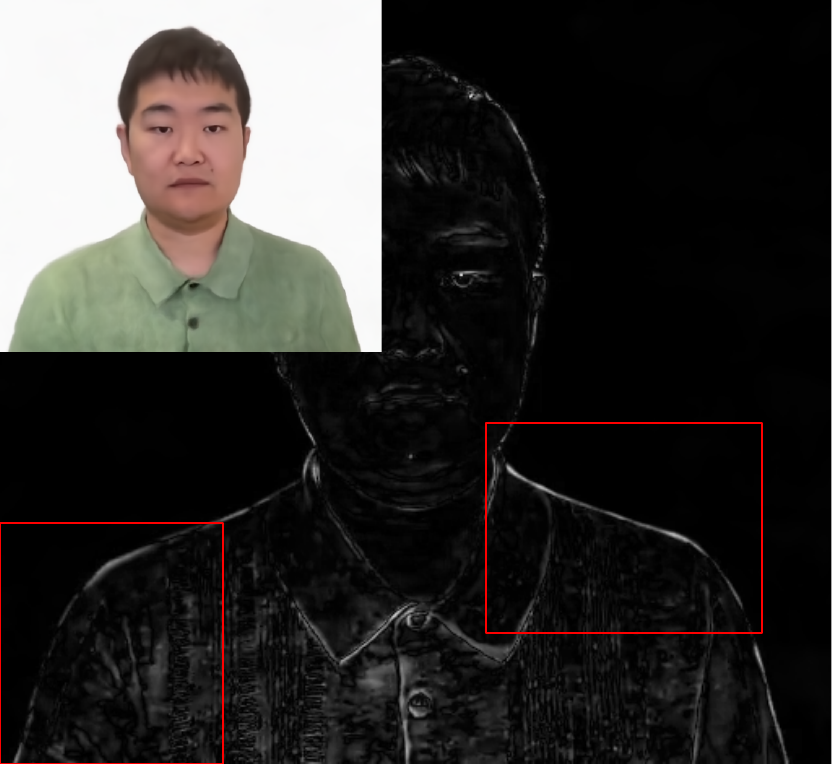} \\
(a) & (b.1) GT & \small (b.2) No warping & \small (b.3) Use warping \\
\end{tabular}
\vspace{-2mm}
\caption{(a) Test LPIPS comparison between using the proposed expression representation (Eq.~\eqref{eq:expressoion_repre}) and the raw expression code in our LightAvatar on \textit{Subject 0}. The two models are trained for the same iterations ($200K$). (b) Comparison between \textit{not using}~and~\textit{using} the proposed \textit{warping field network} (Sec.~\ref{subsec:warping}) in our method. The black-background image refers to the pixel-wise difference between the predicted image and the ground-truth, where \textit{brighter} color indicates \textit{larger} difference.}
\label{fig:ablation_shoulder_joint_vs_separate}
\vspace{-4mm}
\end{figure}

\vspace{-3mm}
\subsection{Results with Shoulders}
\vspace{-2mm}
In Fig.~\ref{fig:results_with_shoulder}(a), we present the results with the shoulders to show the capability of our method in capturing the torso area. Our approach achieves \textit{better} test LPIPS/SSIM/PSNR compared to the teacher MonoAvatar~\cite{bai2023learning}. Visually, our method also produces \textit{sharper} details (such as the textures of the apparel and the reflections in the irises). Notably, for the inside-mouth area (note the yellow arrow), it is not covered by the head mesh. MonoAvatar thus produces more artifacts since it anchors local radiance fields to mesh vertices. In contrast, our mouth area is more visually pleasing since we do not rely on mesh.

In Fig.~\ref{fig:results_with_shoulder}(b), we show the effect of using the proposed \textit{separate modeling} scheme (Sec.~\ref{subsec:lightavatar}) to learn the shoulder area in our LightAvatar. As seen, if the torso rotation is not considered, the shoulder will mistakenly rotate with the head rotation, causing severe miss-alignments and flickering issues. Instead, with the proposed separate modeling scheme, LightAvatar learns to disentangle the 
neck rotation from the rest of the body. As such, the rendered shoulder is not flickering anymore. This shows the encouraging potential of our method to handle different parts of an avatar by a single model.

\subsection{Ablation Study}
\noindent \textbf{Effect of our Expression Representation}. A naive baseline for the expression representation is to use the raw expression code (with positional encoding). Fig.~\ref{fig:ablation_shoulder_joint_vs_separate}(a) presents the comparison between our proposed expression representation and this naive scheme. As seen, our expression representation helps the LightAvatar model converge much faster and achieve a better test LPIPS.

\vspace{0.3em}
\noindent \textbf{Effect of Warping Field Network}. 
Fig.~\ref{fig:ablation_shoulder_joint_vs_separate}(b) shows the comparison of using and not using the warping field network. The warping field network makes the rendered image more aligned with the ground-truth. With the warping network, we can finetune the model on real data, gaining better quality while not undermining the temporal consistency.

%% file: sec/5_conclusion.tex
\vspace{-4mm}
\section{Conclusion and Limitations}
\vspace{-3mm}
\label{sec:conclusion}
This work introduces \textit{LightAvatar}, offering a compelling proof-of-concept for building photo-realistic head avatars with neural \textit{light} fields. LightAvatar features a simple and uniform network design - it takes 3DMM parameters and camera pose as input and predicts the image via a single network forward pass. We introduce dedicated network designs to ensure training stability and high rendering efficiency. The model training is challenging due to the light field formulation and compact network design. To resolve this, we present a \textit{distillation-based} training pipeline and a warp field network so as to mitigate the fitting error in the real data. Extensive results and analyses on many subjects show our LightAvatar reaches SOTA image quality while being significantly faster than the counterparts, rendering at \textbf{174.1 FPS} on a consumer-grade GPU (RTX3090).

\textit{Limitations}. This work still has several limitations to overcome. (1) We do not explicitly consider the more nuanced expressions (such as the eye-ball rotation) in our current framework. The model could learn these variations by itself, but we still observe a few small differences~\textit{vs.}~the ground-truth (Fig.~\ref{fig:limitations}(a) in supp.). Besides, it is also challenging for our method to handle complex structures like long hairs (Fig.~\ref{fig:limitations}(b) in supp.), similar to existing 
top-performing NeRF-based avatars. How to integrate these nuanced details in our framework to make the avatar even more photo-realistic is a worthy next step. (2) This work aimed at building a NeLF-based avatar for faster \textit{inference} in this paper. The \textit{training} efficiency of building the avatars is not improved much in this work (yet still comparable to counterparts such as PointAvatar~\cite{zheng2023pointavatar}). Accelerating training (like INSTA~\cite{zielonka2023instant}) is also worth exploring in the future. (3) 3DGS-based avatars (\eg, GaussianAvatars~\cite{qian2024gaussianavatars}) are competitive approaches for fast rendering. Although our method is much faster than GaussianAvatars~\cite{qian2024gaussianavatars}) ( Tab.~\ref{tab:speed_vs_GA} in supp.), a more comprehensive comparison in terms of the quality is preferred in the future.

%% file: sec/6_supp_eccv24_workshop.tex
\section{Overview of Supplementary Materials}
\label{sec:rationale}
In the following supplementary materials, we provide:
\begin{itemize}
    \item more detailed network architecture explanations  (Sec.~\ref{sec:network_arch});
    \item comparisons with the teacher on three more subjects (Sec.~\ref{subsec:more_subjects});
    \item video comparisons with different methods (Sec.~\ref{subsec:video_results});
    \item speed comparison with a Gaussian-Splatting-based avatar (Sec.~\ref{subsec:gs_avatar});
    \item video comparison between the joint modeling scheme and separate modeling when learning the shoulder (Sec.~\ref{subsec:video_shoulder});
    \item ablation study: visual comparison between using our expression representation~\vs~using the raw expression code as NeLF input (Sec.~\ref{subsec:raw_expr_vs_our_expr}); 
    \item 3D consistency check of our method (Sec.~\ref{subsec:3d_consistency});
    \item potential negative impact discussion (Sec.~\ref{subsec:negative_impact}).
\end{itemize}

\section{Detailed Network Architectures}\label{sec:network_arch}
\begin{figure}[!h]
\centering
\setlength{\tabcolsep}{4mm} %
\begin{tabular}{cccccc}
\includegraphics[trim=0.4cm 0cm 0cm 0cm,clip,width=0.9\linewidth]{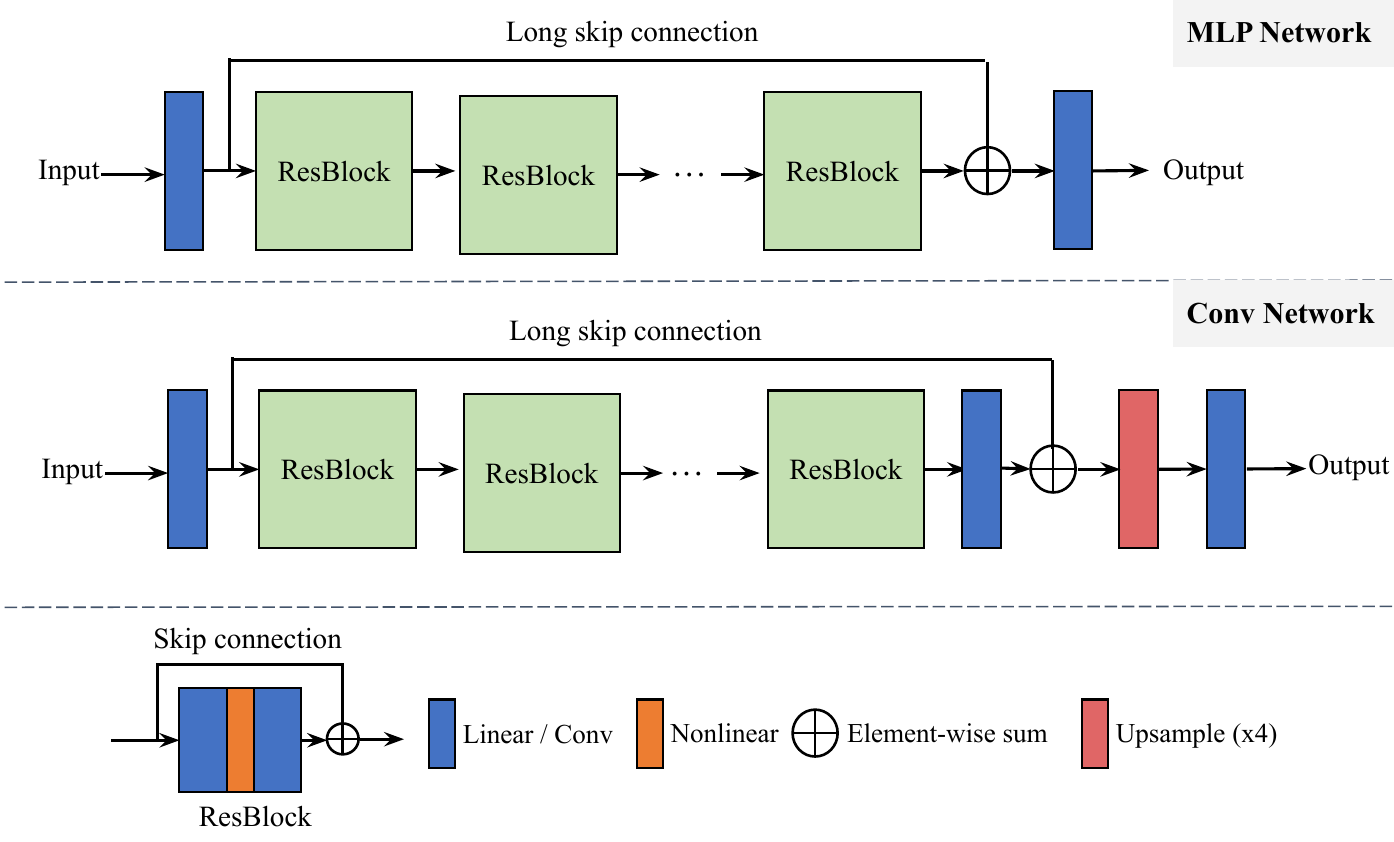} \\
\end{tabular}
\vspace{-2mm}
\caption{Illustration of two types of networks used in our method. We intentionally use simple and uniform network designs for the models in our method - Each model has 3 parts: head, body, and tail, with a long skip connection bypassing the body. }
\label{fig:network_arch}
\end{figure}

Two types of networks are used in our method, MLP network and Conv network. Their architectures, as shown in Fig.~\ref{fig:network_arch}, follow a pretty simple and uniform paradigm, by our particular design.

For the SR model, we use the Conv network architecture; for the others, we use the MLP network architecture. Major differences lie in the number of residual blocks (ResBlocks) used in the body (which determines the network depth), and the width for internal layers. For NeLF network, we use 10 ResBlocks with width 128. For spatial attention network and local feature network, we use 2 ResBlocks with width 128. For SR network, we use 5 ResBlocks with width 56.

LeakyReLU is used for the major non-linearity (unless specified otherwise). We do \textit{not} use any normalization layers (unlike prior related works like MobileR2L~\cite{cao2023real}) because (1) our specially designed training strategy already makes the algorithm converge successfully and fast; (2) in practice, normalization layers will lead to extra inference latency.

\section{Additional Results}
\subsection{Results on \textit{Subject 5} to \textit{Subject 7}} \label{subsec:more_subjects}
The quantitative results of the other three subjects (\textit{Subject 5} to \textit{Subject 7}) are presented in Tab.~\ref{tab:results_quantitative_flame_more_subjects}, visual results in Fig.~\ref{fig:results_visual_flame_more_subjects}. Here we mainly compare with MonoAvatar~\cite{bai2023learning} (the teacher), since it has been shown to be better than the other methods in Tab.~\RE{1}.

As seen, similar to the results (Tab.~\RE{1}, Fig.~\RE{3}) in the main paper, our LightAvatar still consistently achieves \textit{better} quantitative results and slightly better (or comparable) visual quality than the teacher. Here we also observe that our LightAvatar produces less grain noise than MonoAvatar, \eg, the mouth area of \textit{Subject 6}. Note, we achieve these better or comparable results with more than 300$\times$ speedup against MonoAvatar (see Tab.~\RE{3} in the paper). 

For \textit{Subject 5}, both MonoAvatar and our method do not accurately reconstruct the complex long hair structure and extreme facial expression, which is considered as a limitation of our method now (as discussed in the paper, Sec.~\RE{5}).

\begin{table*}[!h]
\centering
\caption{LPIPS$\downarrow$/SSIM$\uparrow$/PSNR$\uparrow$ comparison on \textit{Subject 5/6/7}. Here we compare our method to the teacher MonoAvatar~\cite{bai2023learning}. As seen, similar to the results (Tab.~\RE{1}) in the main paper, our method is consistently better than teacher. Note, we achieve so meanwhile being more than 300$\times$ faster than MonoAvatar (Tab.~\RE{3} in the main paper).}
\vspace{-2mm}
\resizebox{0.99\linewidth}{!}{
\setlength{\tabcolsep}{2mm}
\begin{tabular}{l|c|c|c|c}
\multirow{2}{*}{Method} & \textit{Subject 5} & \textit{Subject 6} & \textit{Subject 7} & \textit{Average} \\
& LPIPS/SSIM/PSNR & LPIPS/SSIM/PSNR & LPIPS/SSIM/PSNR & LPIPS/SSIM/PSNR \\
\Xhline{3\arrayrulewidth}
MonoAvatar & 0.246/0.704/16.57 & 0.160/0.862/21.70 & 0.119/0.891/23.45 & 0.175/0.819/20.57 \\
\rowcolor[gray]{0.92} Ours & 0.231/0.709/16.78 & 0.147/0.864/21.97 & 0.102/0.894/23.70 & \textbf{0.160}/\textbf{0.822}/\textbf{20.82} \\
\end{tabular}}
\label{tab:results_quantitative_flame_more_subjects}
\end{table*}

\begin{figure}[!h]
\centering
\resizebox{0.8\linewidth}{!}{
\setlength{\tabcolsep}{4mm} %
\begin{tabular}{cccccc}
\includegraphics[trim=1cm 1cm 0cm 1cm,clip,width=0.32\linewidth]{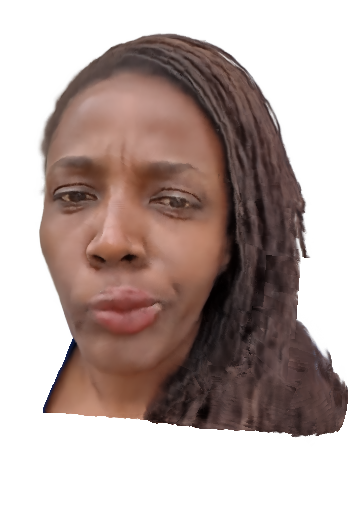} &
\includegraphics[trim=1cm 1cm 0cm 1cm,clip,width=0.32\linewidth]{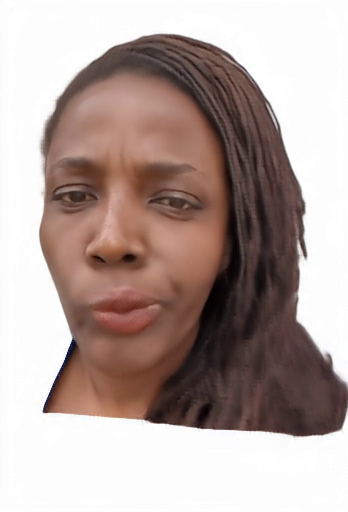} &
\includegraphics[trim=1cm 1cm 0cm 1cm,clip,width=0.32\linewidth]{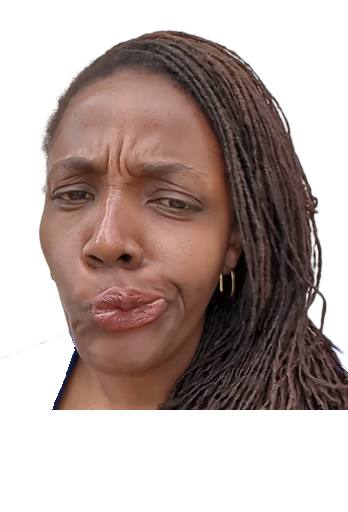} \\
\includegraphics[trim=1cm 1cm 1cm 1cm,clip,width=0.32\linewidth]{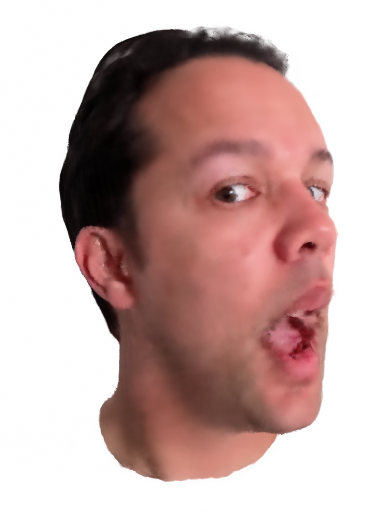} &
\includegraphics[trim=1cm 1cm 1cm 1cm,clip,width=0.32\linewidth]{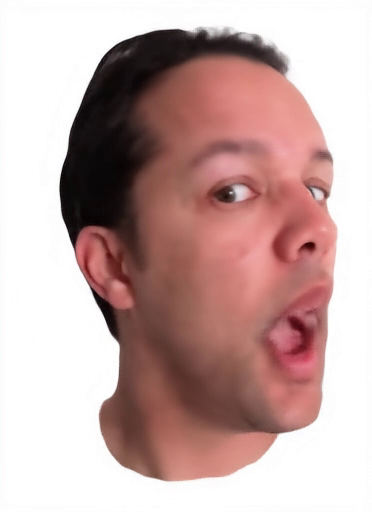} &
\includegraphics[trim=1cm 1cm 1cm 1cm,clip,width=0.32\linewidth]{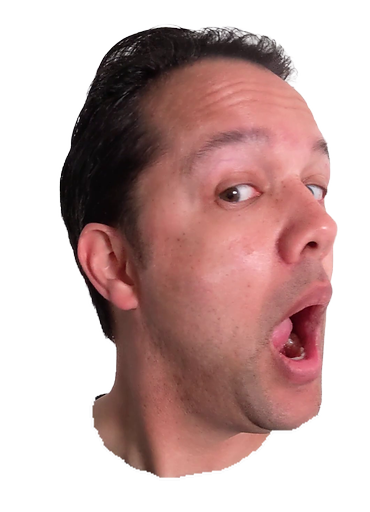} \\
\includegraphics[trim=1cm 1cm 1cm 0cm,clip,width=0.32\linewidth]{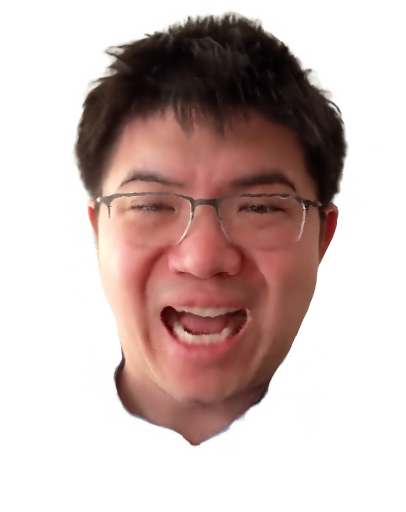} &
\includegraphics[trim=1cm 1cm 1cm 0cm,clip,width=0.32\linewidth]{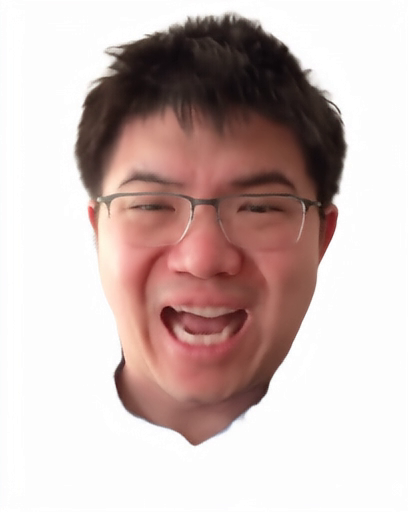} &
\includegraphics[trim=1cm 1cm 1cm 0cm,clip,width=0.32\linewidth]{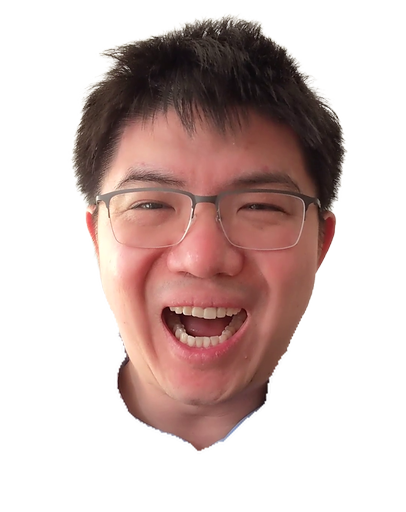} \\
(a) MonoAvatar & (b) Ours & (c) GT \\
\end{tabular}}
\vspace{-2mm}
\caption{Visual comparison with MonoAvatar~\cite{bai2023learning} on the test set. From top to down, the subject is \textit{Subject5} to \textit{Subject7} in order.}
\label{fig:results_visual_flame_more_subjects}
\vspace{-4mm}
\end{figure}

\subsection{Limitations of Our Method}
\begin{figure}[!h]
\centering
\setlength{\tabcolsep}{1mm}
\begin{tabular}{cccccc}
\includegraphics[trim=1cm 1cm 2cm 1cm,clip,width=0.199\linewidth]{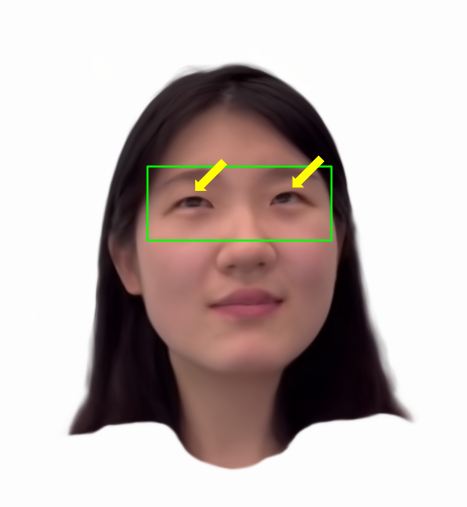} &
\includegraphics[trim=1cm 1cm 2cm 1cm,clip,width=0.199\linewidth]{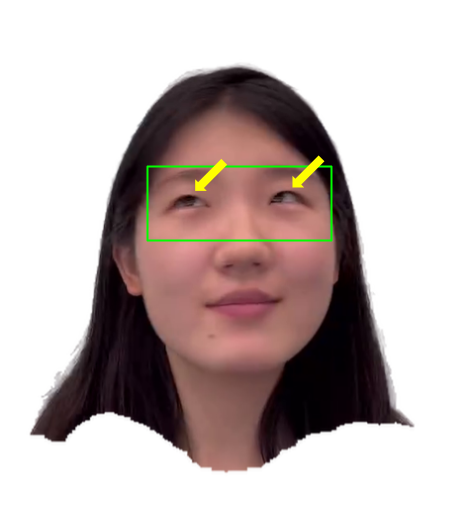} &
\includegraphics[trim=1cm 2cm 0cm 1cm,clip,width=0.165\linewidth]{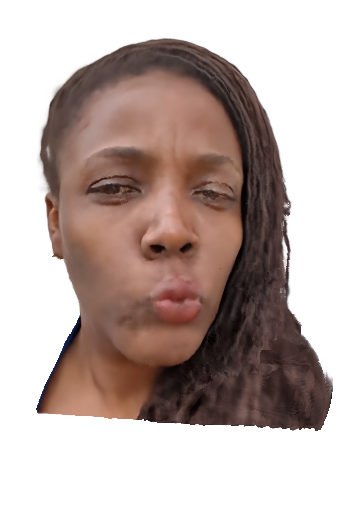} &
\includegraphics[trim=1cm 2cm 0cm 1cm,clip,width=0.165\linewidth]{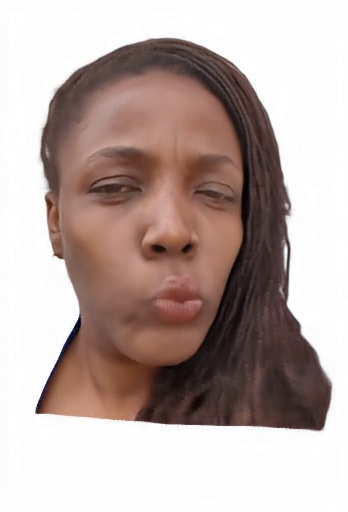} &
\includegraphics[trim=1cm 2cm 0cm 1cm,clip,width=0.165\linewidth]{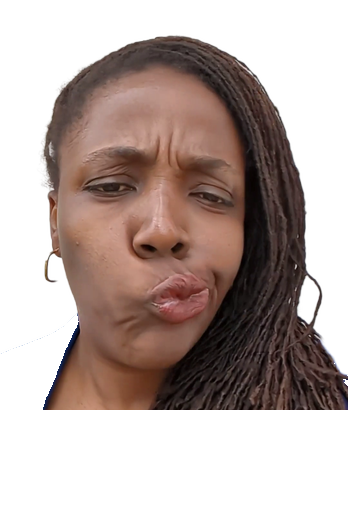} \\
(a.1) Ours & (a.2) GT & (b.1) MonoAvatar & (b.2) Ours & (b.3) GT  
\end{tabular}
\vspace{-2mm}
\caption{Showcases of limitations of our method. (a) Our method does not explicitly integrate the eyeball rotation into the framework at present, so it is still challenging for our method to learn the correct eyeball rotation implicitly. (b) For the complex structures, such as long hairs, our method still cannot accurately reconstruct the details, similar to the NeRF-based SOTA methods (\eg, MonoAvatar~\cite{bai2023learning}).}
\vspace{-5mm}
\label{fig:limitations}
\end{figure}

Fig.~\ref{fig:limitations} showcases two limitations of our method: It is challenging to capture the nuanced details and model complex structures like long hairs.

\subsection{Video Comparison w. Other Methods}\label{subsec:video_results}
In the paper, we presented two sets of comparisons of our method~\vs~the prior SOTA avatars (Tab.~\RE{1}, Fig.~\RE{3}) and the recent fast avatars specialized for fast inference speed (Tab.~\RE{2}, Fig.~\RE{4}). The corresponding videos are presented in the attached webpage \BL{\texttt{index.html}} (see ``\textit{1.~Video Comparison with 5 Subjects}'', and ``\textit{2.~Video Comparison with Fast Avatars}''). Please find the videos in our code repository:~\href{https://github.com/MingSun-Tse/LightAvatar-TensorFlow}{https://github.com/MingSun-Tse/LightAvatar-TensorFlow}.

\subsection{Speed Comparison w.~3DGS-Based Avatar}\label{subsec:gs_avatar}
Recently, 3D \textit{Gaussian Splatting} (3DGS)~\cite{kerbl20233d} has become a new emerging 3D representation, which has been successfully extended to building photo-realistic head avatars, such as GaussianAvatar~\cite{qian2024gaussianavatars}.

Our work fundamentally differs from these GS-based avatars since we build upon \textit{light fields} instead of Gaussians. GS is also known for its fast rendering speed~\cite{kerbl20233d} based on rasterization. GaussianAvatar inherits this advantage. The rendering speed benchmarks in Tab.~\ref{tab:speed_vs_GA} show that LightAvatar runs more slowly as the resolution goes higher (since the computation becomes proportionally larger \textit{w.r.t.} output resolution), while GaussianAvatar keeps a nearly constant speed (since GA uses rasterization for rendering, which has been highly optimized on modern GPUs; the rendering speed is barely bound by the output resolution). As a result, at small resolutions, LightAvatar is faster; at larger resolutions, GaussianAvatar is faster. Simply put, LightAvatar gets the fast speed by \textit{algorithm designs}; GaussianAvatar gets the fast speed by \textit{hardware support} - clearly, they follow different paths. We envision one day when neural operators are as optimized as rasterization on GPUs, the potential of our method can be more fulfilled.

\begin{table*}[!h]
\centering
\caption{Speed (FPS) comparison with GaussianAvatar (GA)~\cite{qian2024gaussianavatars} across different output resolutions, on an RTX4090 GPU (24GB) with PyTorch~\cite{pytorch}. We use the public code$^\dagger$ of GA for this benchmark, speed averaged by 500 frames following the guidelines of GA.}
\vspace{-2mm}
\resizebox{0.95\linewidth}{!}{
\setlength{\tabcolsep}{3mm}
\begin{tabular}{l|c|c|c|c|c|c|cc}
Resolution & 128x128 & 256x256 & 384x384 & 512x512 & 640x640 & 768x768 & 896x896 \\
\Xhline{3\arrayrulewidth}
GA & 189.1 & 201.0 & 211.3 & \textbf{217.1} & \textbf{215.8} & \textbf{212.0} & \textbf{210.8} \\
\rowcolor[gray]{0.92} Ours & \textbf{408.1} & \textbf{331.4} & \textbf{295.1} & 186.3 & 119.0 & 82.3 & 62.3 \\
\end{tabular}}
\begin{tablenotes}
    \item \footnotesize{\texttt{$^\dagger$https://github.com/ShenhanQian/GaussianAvatars}}
\end{tablenotes}
\label{tab:speed_vs_GA}
\end{table*}

\subsection{Video Comparison of Joint~\vs~Separate Modeling of Shoulder}\label{subsec:video_shoulder}
The video comparison is shown in the webpage \BL{\texttt{index.html}} (please refer to the section ``\textit{3.~Video comparison: Joint~\vs~separate modeling in our method}'').

Clearly, the proposed separate modeling scheme achieves \textit{much better} temporal consistency than the joint modeling, showing the encouraging potential of our method learning the head and torso with a \textit{single} model.

\subsection{Visual Comparison of Different Expression Representations} \label{subsec:raw_expr_vs_our_expr}
\begin{figure}[!h]
\centering
\setlength{\tabcolsep}{1mm} %
\begin{tabular}{cccccc}
\includegraphics[width=0.32\linewidth]{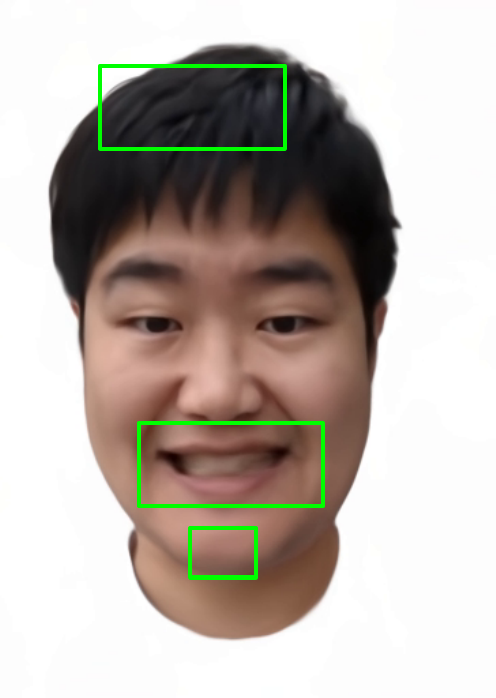} &
\includegraphics[width=0.32\linewidth]{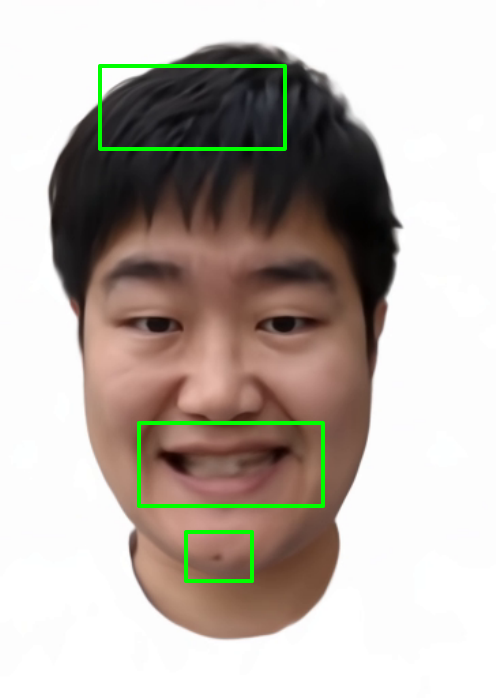} &
\includegraphics[width=0.32\linewidth]{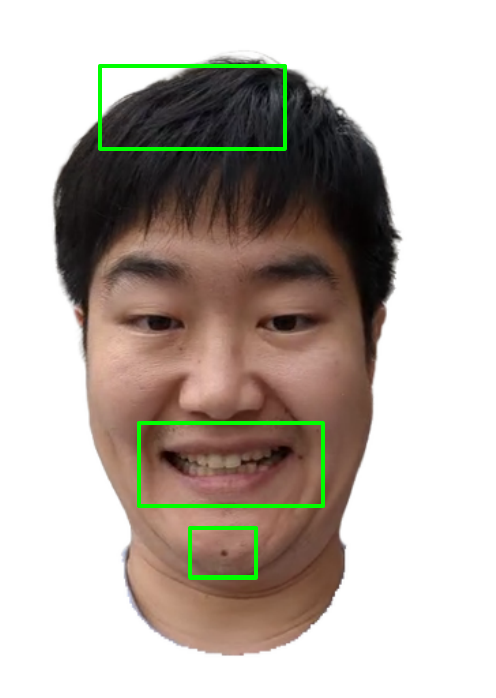} \\
(a) Raw expression & (b) Our expression & (c) GT
\end{tabular}
\vspace{-2mm}
\caption{Comparison between using raw expression code (a)~\vs~our expression representation (b) for the NeLF backbone in our LightAvatar (on \textit{Subject 0}).}
\label{fig:raw_expr_vs_our_expr}
\vspace{-4mm}
\end{figure}

One of the key contribution claimed in the paper is the designed expression representation (\textit{Sec.~\RE{3.2}, (2) Expression Representation}) ~\vs~using the raw expression code as input to regress the target RGB. In the paper, we presented the evidence in terms of the test LPIPS (Fig.~\RE{6}(a)) to show the merit of our expression representation. Here we present the visual comparison in Fig.~\ref{fig:raw_expr_vs_our_expr} to further validate our claim -- Using our expression helps reconstruct clearly sharper textures.

\subsection{Visual Comparison of Different Expression Representations} \label{subsec:3d_consistency}
\begin{figure}[!h]
\centering
\resizebox{1\linewidth}{!}{
\setlength{\tabcolsep}{0.1mm} %
\begin{tabular}{ccccccc}
\includegraphics[trim=30cm 16cm 30cm 20cm,clip,height=0.18\linewidth]{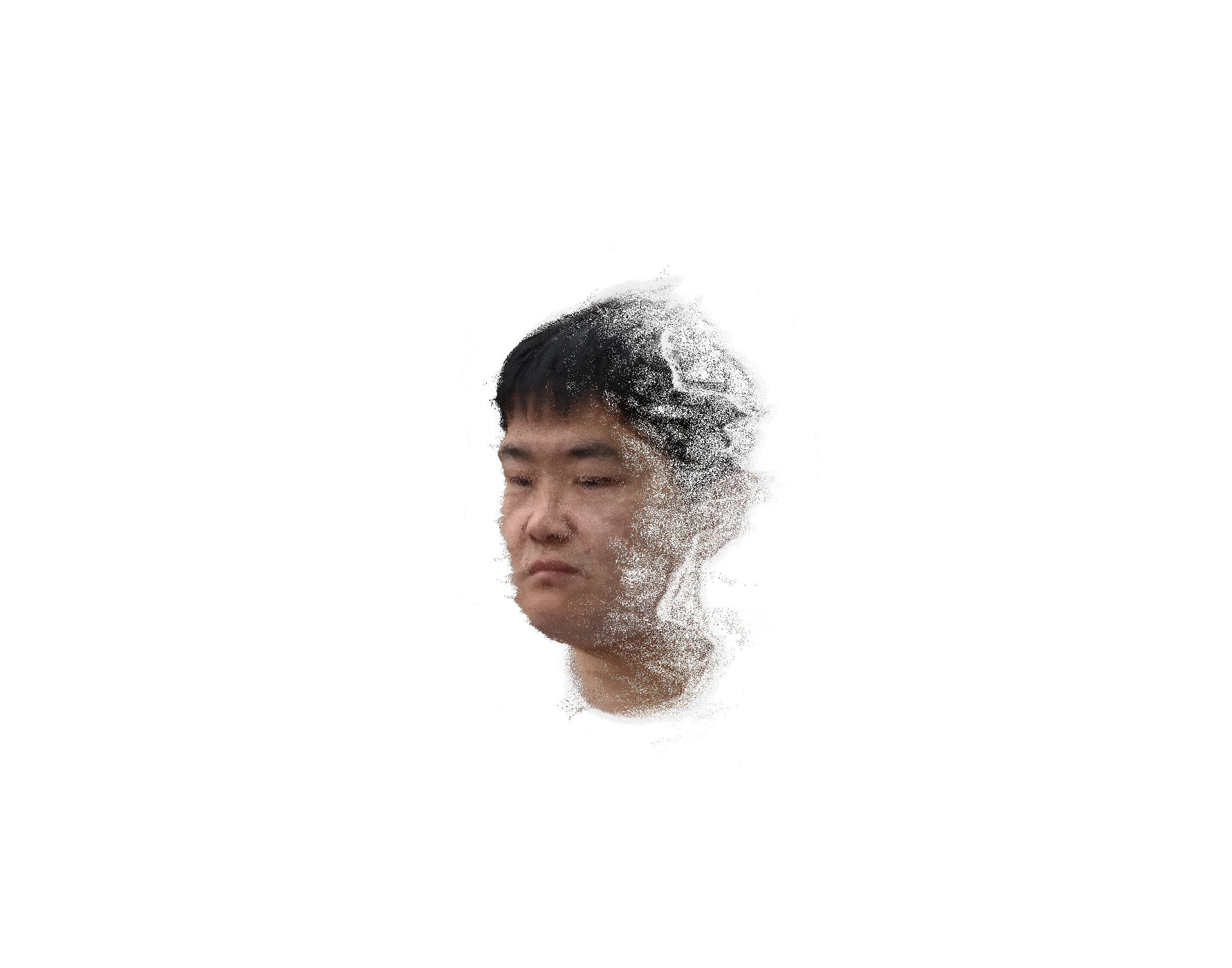} &
\includegraphics[trim=30cm 16cm 30cm 20cm,clip,height=0.18\linewidth]{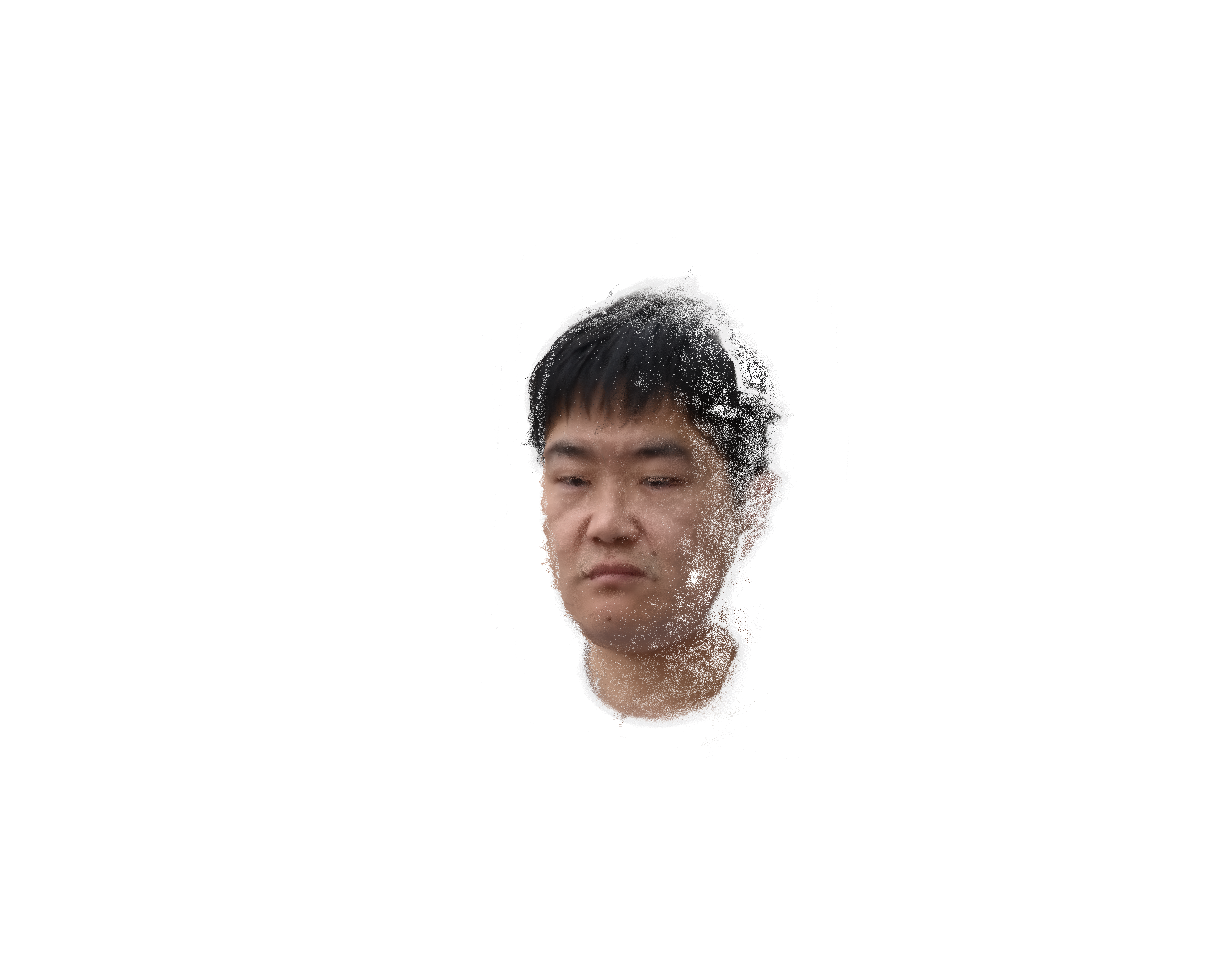} &
\includegraphics[trim=30cm 16cm 30cm 20cm,clip,height=0.18\linewidth]{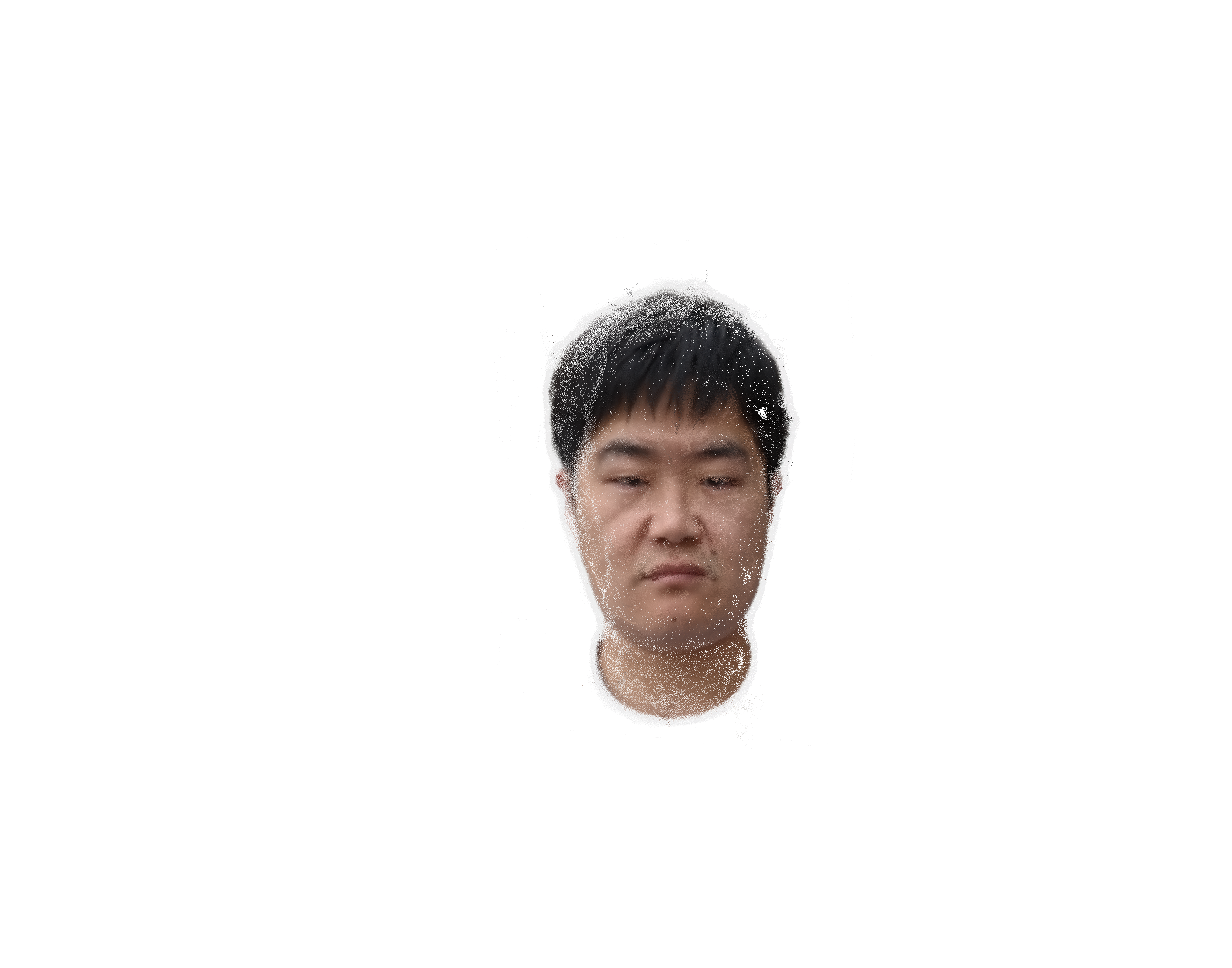} &
\includegraphics[trim=30cm 16cm 30cm 20cm,clip,height=0.18\linewidth]{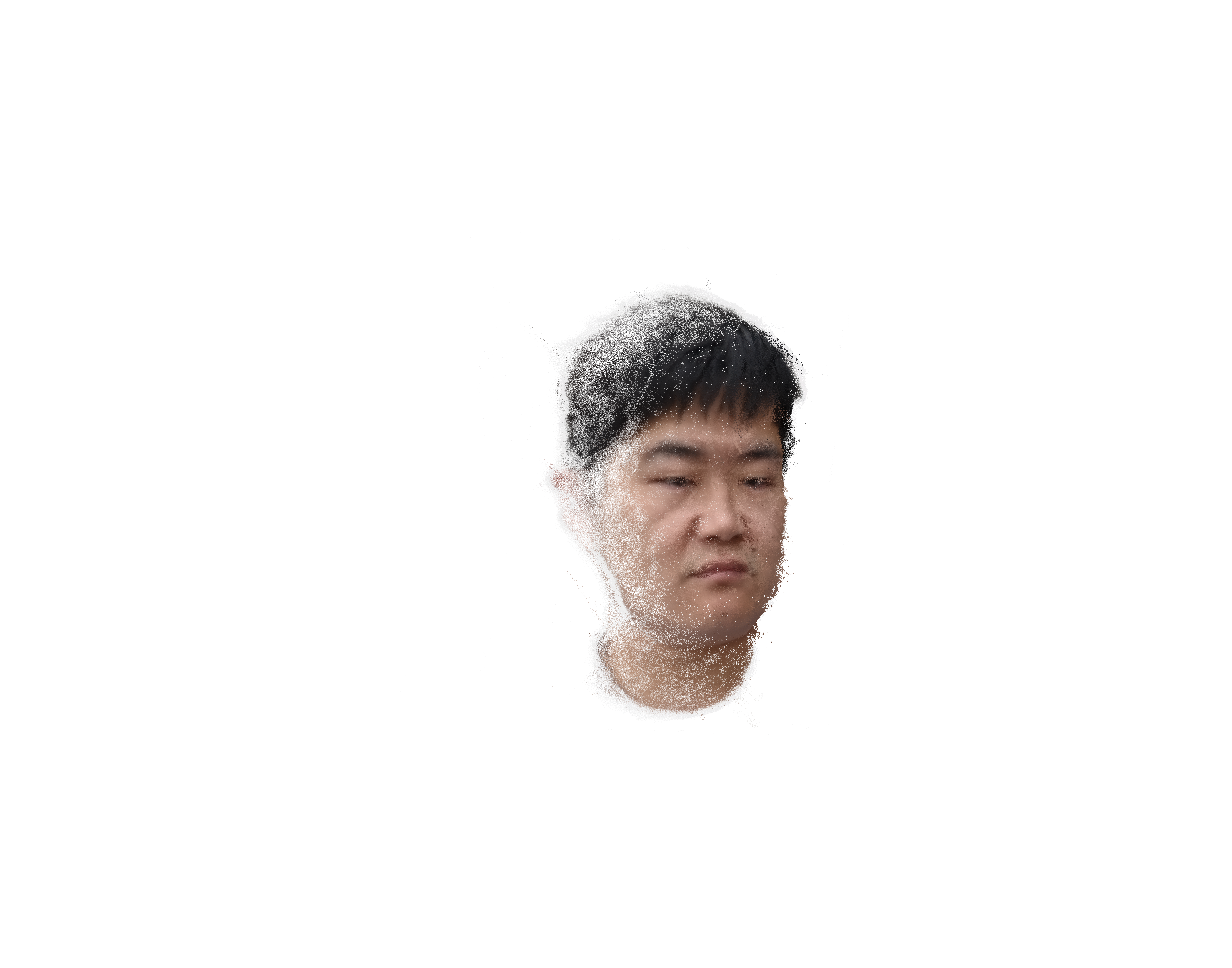} &
\includegraphics[trim=30cm 16cm 30cm 20cm,clip,height=0.18\linewidth]{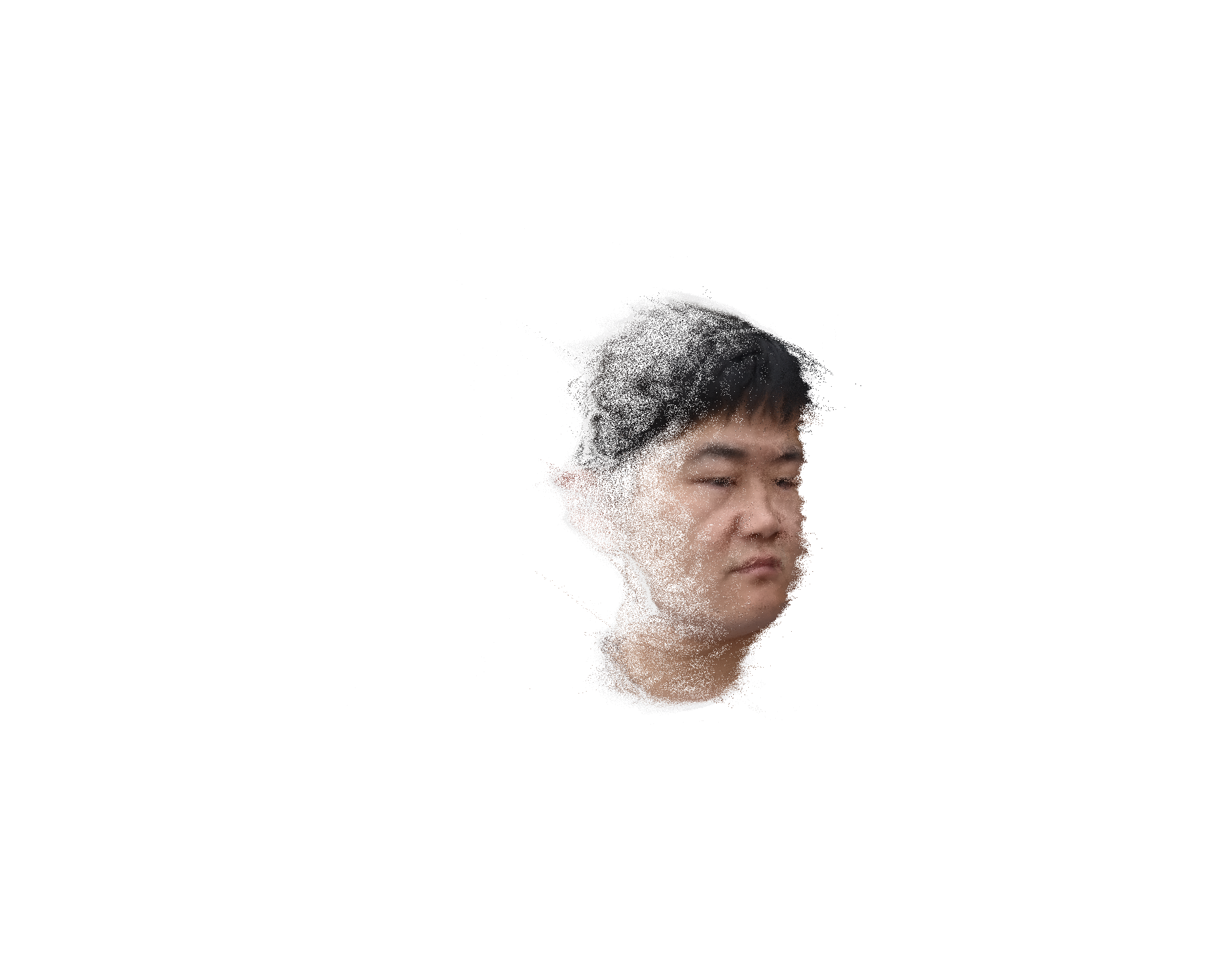} &
\end{tabular}}
\vspace{-2mm}
\caption{3D point cloud reconstruction with COLMAP. As seen, our method can reconstruct consistent 3D shapes despite not using any explicit 3D designs.}
\label{fig:colmap}
\end{figure}

Fig.~\ref{fig:colmap} shows the reconstructed point clouds by COLMAP~\cite{schoenberger2016sfm,schoenberger2016mvs} with camera poses of the test set 434 images of \textit{Subject 0}. As seen, although our method does not has any explicit constraint for 3D consistency (unlike NeRFs), it \textit{learns by itself} the view consistency, thanks to the massive (20K) pseudo images and sufficient capacity. This agrees with the observation in prior static-scene NeLFs~\cite{wang2022r2l,cao2023real}.

\subsection{Potential Negative Impact}\label{subsec:negative_impact}
This work aims to build photo-realistic head avatars, which can be used in negative technology such as ``Deep Fake''~\cite{nguyen2022deep}. This has been a general potential issue for all avatar methods. To prevent misuse, we need to ensure the right access control. More advanced security measures need to be developed at the same time as avatars become more photo-realistic, such as encryption and secure storage of the avatars, using watermarks and digital signatures.

Besides, in some contexts, \eg, social VR environments, the use of head avatars could raise privacy concerns if personal information or sensitive data is collected or exposed without users' consent. To prevent this, ethical guidelines and collaborative efforts from the society are indispensable.

%% file: main.bbl
\begin{thebibliography}{10}
\providecommand{\url}[1]{\texttt{#1}}
\providecommand{\urlprefix}{URL }
\providecommand{\doi}[1]{https://doi.org/#1}

\bibitem{abadi2016tensorflow}
Abadi, M., Barham, P., Chen, J., Chen, Z., Davis, A., Dean, J., Devin, M., Ghemawat, S., Irving, G., Isard, M., et~al.: {TensorFlow}: A system for large-scale machine learning. In: IEEE Symposium on Operating Systems Design and Implementation (2016)

\bibitem{athar2022rignerf}
Athar, S., Xu, Z., Sunkavalli, K., Shechtman, E., Shu, Z.: Rignerf: Fully controllable neural 3d portraits. In: {CVPR} 2022. pp. 20332--20341 (2022)

\bibitem{attal2022learning}
Attal, B., Huang, J.B., Zollh{\"o}fer, M., Kopf, J., Kim, C.: Learning neural light fields with ray-space embedding. In: CVPR (2022)

\bibitem{bai2021riggable}
Bai, Z., Cui, Z., Liu, X., Tan, P.: Riggable 3d face reconstruction via in-network optimization. In: {CVPR} 2021. pp. 6216--6225 (2021)

\bibitem{bai2020deep}
Bai, Z., Cui, Z., Rahim, J.A., Liu, X., Tan, P.: Deep facial non-rigid multi-view stereo. In: {CVPR} 2020. pp. 5849--5859 (2020)

\bibitem{bai2023learning}
Bai, Z., Tan, F., Huang, Z., Sarkar, K., Tang, D., Qiu, D., Meka, A., Du, R., Dou, M., Orts-Escolano, S., et~al.: Learning personalized high quality volumetric head avatars from monocular rgb videos. In: CVPR (2023)

\bibitem{barron2021mip}
Barron, J.T., Mildenhall, B., Tancik, M., Hedman, P., Martin-Brualla, R., Srinivasan, P.P.: Mip-nerf: A multiscale representation for anti-aliasing neural radiance fields. In: ICCV (2021)

\bibitem{barron2022mip}
Barron, J.T., Mildenhall, B., Verbin, D., Srinivasan, P.P., Hedman, P.: Mip-nerf 360: Unbounded anti-aliased neural radiance fields. In: CVPR (2022)

\bibitem{bemana2020x}
Bemana, M., Myszkowski, K., Seidel, H.P., Ritschel, T.: X-fields: Implicit neural view-, light-and time-image interpolation. ACM Transactions on Graphics  \textbf{39}(6),  1--15 (2020)

\bibitem{bharadwaj2023flare}
Bharadwaj, S., Zheng, Y., Hilliges, O., Black, M.J., Fernandez-Abrevaya, V.: Flare: Fast learning of animatable and relightable mesh avatars. In: SIGGRAPH Asia (2023)

\bibitem{blanz1999morphable}
Blanz, V., Vetter, T.: A morphable model for the synthesis of 3d faces. In: SIGGRAPH (1999)

\bibitem{bucilua2006model}
Buciluǎ, C., Caruana, R., Niculescu-Mizil, A.: Model compression. In: SIGKDD (2006)

\bibitem{cao2016real}
Cao, C., Wu, H., Weng, Y., Shao, T., Zhou, K.: Real-time facial animation with image-based dynamic avatars. {ACM} Trans. Graph.  \textbf{35}(4),  126:1--126:12 (2016)

\bibitem{cao2023real}
Cao, J., Wang, H., Chemerys, P., Shakhrai, V., Hu, J., Fu, Y., Makoviichuk, D., Tulyakov, S., Ren, J.: Real-time neural light field on mobile devices. In: CVPR (2023)

\bibitem{chaudhuri2020personalized}
Chaudhuri, B., Vesdapunt, N., Shapiro, L.G., Wang, B.: Personalized face modeling for improved face reconstruction and motion retargeting. In: {ECCV} 2020. vol. 12350, pp. 142--160 (2020)

\bibitem{chen2022tensorf}
Chen, A., Xu, Z., Geiger, A., Yu, J., Su, H.: Tensorf: Tensorial radiance fields. In: {ECCV} 2022. vol. 13692, pp. 333--350 (2022)

\bibitem{chen2023mobilenerf}
Chen, Z., Funkhouser, T.A., Hedman, P., Tagliasacchi, A.: Mobilenerf: Exploiting the polygon rasterization pipeline for efficient neural field rendering on mobile architectures. In: {CVPR} 2023. pp. 16569--16578. {IEEE} (2023)

\bibitem{chen2019learning}
Chen, Z., Zhang, H.: Learning implicit fields for generative shape modeling. In: CVPR (2019)

\bibitem{egger20203d}
Egger, B., Smith, W.A., Tewari, A., Wuhrer, S., Zollhoefer, M., Beeler, T., Bernard, F., Bolkart, T., Kortylewski, A., Romdhani, S., et~al.: 3d morphable face models—past, present, and future. ACM Trans. Graphic.  \textbf{39}(5),  1--38 (2020)

\bibitem{gafni2021dynamic}
Gafni, G., Thies, J., Zollhofer, M., Nie{\ss}ner, M.: Dynamic neural radiance fields for monocular 4d facial avatar reconstruction. In: CVPR (2021)

\bibitem{gao2022reconstructing}
Gao, X., Zhong, C., Xiang, J., Hong, Y., Guo, Y., Zhang, J.: Reconstructing personalized semantic facial nerf models from monocular video. {ACM} Trans. Graph.  \textbf{41}(6),  200:1--200:12 (2022)

\bibitem{garbin2021fastnerf}
Garbin, S.J., Kowalski, M., Johnson, M., Shotton, J., Valentin, J.: Fastnerf: High-fidelity neural rendering at 200fps. arXiv preprint arXiv:2103.10380  (2021)

\bibitem{garrido2014automatic}
Garrido, P., Valgaerts, L., Rehmsen, O., Thorm{\"{a}}hlen, T., P{\'{e}}rez, P., Theobalt, C.: Automatic face reenactment. In: {CVPR} 2014. pp. 4217--4224 (2014)

\bibitem{garrido2016reconstruction}
Garrido, P., Zollh{\"{o}}fer, M., Casas, D., Valgaerts, L., Varanasi, K., P{\'{e}}rez, P., Theobalt, C.: Reconstruction of personalized 3d face rigs from monocular video. {ACM} Trans. Graph.  \textbf{35}(3),  28:1--28:15 (2016)

\bibitem{gortler1996lumigraph}
Gortler, S.J., Grzeszczuk, R., Szeliski, R., Cohen, M.F.: The lumigraph. In: Proceedings of the Annual Conference on Computer Graphics and Interactive Techniques (1996)

\bibitem{grassal2022neural}
Grassal, P.W., Prinzler, M., Leistner, T., Rother, C., Nie{\ss}ner, M., Thies, J.: Neural head avatars from monocular rgb videos. In: CVPR (2022)

\bibitem{gupta2023lightspeed}
Gupta, A., Cao, J., Wang, C., Hu, J., Tulyakov, S., Ren, J., Jeni, L.: Lightspeed: light and fast neural light fields on mobile devices. In: NeurIPS (2023)

\bibitem{hedman2021baking}
Hedman, P., Srinivasan, P.P., Mildenhall, B., Barron, J.T., Debevec, P.: Baking neural radiance fields for real-time view synthesis. In: ICCV (2021)

\bibitem{hinton2015distilling}
Hinton, G., Vinyals, O., Dean, J.: Distilling the knowledge in a neural network. In: NeurIPS Workshop (2014)

\bibitem{hu2017avatar}
Hu, L., Saito, S., Wei, L., Nagano, K., Seo, J., Fursund, J., Sadeghi, I., Sun, C., Chen, Y., Li, H.: Avatar digitization from a single image for real-time rendering. {ACM} Trans. Graph.  \textbf{36}(6),  195:1--195:14 (2017)

\bibitem{ichim2015dynamic}
Ichim, A.E., Bouaziz, S., Pauly, M.: Dynamic 3d avatar creation from hand-held video input. {ACM} Trans. Graph.  \textbf{34}(4),  45:1--45:14 (2015)

\bibitem{jiang2022neuman}
Jiang, W., Yi, K.M., Samei, G., Tuzel, O., Ranjan, A.: Neuman: Neural human radiance field from a single video. In: ECCV (2022)

\bibitem{johnson2016perceptual}
Johnson, J., Alahi, A., Fei-Fei, L.: Perceptual losses for real-time style transfer and super-resolution. In: ECCV (2016)

\bibitem{kajiya1984ray}
Kajiya, J.T., Von~Herzen, B.P.: Ray tracing volume densities. SIGGRAPH  \textbf{18}(3),  165--174 (1984)

\bibitem{kalantari2016learning}
Kalantari, N.K., Wang, T.C., Ramamoorthi, R.: Learning-based view synthesis for light field cameras. ACM Transactions on Graphics  \textbf{35}(6),  1--10 (2016)

\bibitem{karis2013real}
Karis, B., Games, E.: Real shading in unreal engine 4. Proc. Physically Based Shading Theory Practice  \textbf{4}(3), ~1 (2013)

\bibitem{kerbl20233d}
Kerbl, B., Kopanas, G., Leimk{\"u}hler, T., Drettakis, G.: 3d gaussian splatting for real-time radiance field rendering. ACM ToG  \textbf{42}(4),  1--14 (2023)

\bibitem{kim2018deep}
Kim, H., Garrido, P., Tewari, A., Xu, W., Thies, J., Nie{\ss}ner, M., P{\'{e}}rez, P., Richardt, C., Zollh{\"{o}}fer, M., Theobalt, C.: Deep video portraits. {ACM} Trans. Graph.  \textbf{37}(4), ~163 (2018)

\bibitem{kingma2014adam}
Kingma, D.P., Ba, J.: Adam: A method for stochastic optimization. In: ICLR (2015)

\bibitem{laine2020modular}
Laine, S., Hellsten, J., Karras, T., Seol, Y., Lehtinen, J., Aila, T.: Modular primitives for high-performance differentiable rendering. ToG  \textbf{39}(6),  1--14 (2020)

\bibitem{levoy1996light}
Levoy, M., Hanrahan, P.: Light field rendering. In: Proceedings of the Annual Conference on Computer Graphics and Interactive Techniques (1996)

\bibitem{li2017learning}
Li, T., Bolkart, T., Black, M.J., Li, H., Romero, J.: Learning a model of facial shape and expression from 4d scans. In: SIGGRAPH Asia (2017)

\bibitem{lim2017enhanced}
Lim, B., Son, S., Kim, H., Nah, S., Mu~Lee, K.: Enhanced deep residual networks for single image super-resolution. In: CVPR Workshop (2017)

\bibitem{lindell2021autoint}
Lindell, D.B., Martel, J.N., Wetzstein, G.: Autoint: Automatic integration for fast neural volume rendering. In: CVPR (2021)

\bibitem{liu2022neulf}
Liu, C., Li, Z., Yuan, J., Xu, Y.: Neulf: Efficient novel view synthesis with neural 4d light field. In: EGSR (2022)

\bibitem{ma2021pixel}
Ma, S., Simon, T., Saragih, J., Wang, D., Li, Y., De~La~Torre, F., Sheikh, Y.: Pixel codec avatars. In: CVPR (2021)

\bibitem{max1995optical}
Max, N.: Optical models for direct volume rendering. TVCG  \textbf{1}(2),  99--108 (1995)

\bibitem{mescheder2019occupancy}
Mescheder, L., Oechsle, M., Niemeyer, M., Nowozin, S., Geiger, A.: Occupancy networks: Learning 3d reconstruction in function space. In: CVPR (2019)

\bibitem{mildenhall2019local}
Mildenhall, B., Srinivasan, P.P., Ortiz-Cayon, R., Kalantari, N.K., Ramamoorthi, R., Ng, R., Kar, A.: Local light field fusion: Practical view synthesis with prescriptive sampling guidelines. ACM Transactions on Graphics  \textbf{38}(4),  1--14 (2019)

\bibitem{mildenhall2020nerf}
Mildenhall, B., Srinivasan, P.P., Tancik, M., Barron, J.T., Ramamoorthi, R., Ng, R.: Nerf: Representing scenes as neural radiance fields for view synthesis. In: ECCV (2020)

\bibitem{muller2022instant}
M{\"{u}}ller, T., Evans, A., Schied, C., Keller, A.: Instant neural graphics primitives with a multiresolution hash encoding. {ACM} Trans. Graph.  \textbf{41}(4),  102:1--102:15 (2022)

\bibitem{neff2021donerf}
Neff, T., Stadlbauer, P., Parger, M., Kurz, A., Mueller, J.H., Chaitanya, C.R.A., Kaplanyan, A.S., Steinberger, M.: {DONeRF: Towards Real-Time Rendering of Compact Neural Radiance Fields using Depth Oracle Networks}. Computer Graphics Forum  (2021)

\bibitem{nguyen2022deep}
Nguyen, T.T., Nguyen, Q.V.H., Nguyen, D.T., Nguyen, D.T., Huynh-The, T., Nahavandi, S., Nguyen, T.T., Pham, Q.V., Nguyen, C.M.: Deep learning for deepfakes creation and detection: A survey. CVIU  \textbf{223},  103525 (2022)

\bibitem{park2019deepsdf}
Park, J.J., Florence, P., Straub, J., Newcombe, R., Lovegrove, S.: Deepsdf: Learning continuous signed distance functions for shape representation. In: CVPR (2019)

\bibitem{pytorch}
Paszke, A., Gross, S., Massa, F., Lerer, A., Bradbury, J., Chanan, G., Killeen, T., Lin, Z., Gimelshein, N., Antiga, L., et~al.: Pytorch: An imperative style, high-performance deep learning library. In: NeurIPS (2019)

\bibitem{qian2024gaussianavatars}
Qian, S., Kirschstein, T., Schoneveld, L., Davoli, D., Giebenhain, S., Nie{\ss}ner, M.: Gaussianavatars: Photorealistic head avatars with rigged 3d gaussians. In: CVPR (2024)

\bibitem{rebain2021derf}
Rebain, D., Jiang, W., Yazdani, S., Li, K., Yi, K.M., Tagliasacchi, A.: Derf: Decomposed radiance fields. In: CVPR (2021)

\bibitem{reiser2021kilonerf}
Reiser, C., Peng, S., Liao, Y., Geiger, A.: Kilonerf: Speeding up neural radiance fields with thousands of tiny mlps. In: ICCV (2021)

\bibitem{schoenberger2016sfm}
Sch\"{o}nberger, J.L., Frahm, J.M.: Structure-from-motion revisited. In: CVPR (2016)

\bibitem{schoenberger2016mvs}
Sch\"{o}nberger, J.L., Zheng, E., Pollefeys, M., Frahm, J.M.: Pixelwise view selection for unstructured multi-view stereo. In: ECCV (2016)

\bibitem{siarohin2019first}
Siarohin, A., Lathuili{\`e}re, S., Tulyakov, S., Ricci, E., Sebe, N.: First order motion model for image animation. In: NeurIPS (2019)

\bibitem{Simonyan2014Very}
Simonyan, K., Zisserman, A.: Very deep convolutional networks for large-scale image recognition. In: ICLR (2015)

\bibitem{sitzmann2021light}
Sitzmann, V., Rezchikov, S., Freeman, W.T., Tenenbaum, J.B., Durand, F.: Light field networks: Neural scene representations with single-evaluation rendering. In: NeurIPS (2021)

\bibitem{suhail2022light}
Suhail, M., Esteves, C., Sigal, L., Makadia, A.: Light field neural rendering. In: CVPR (2022)

\bibitem{sun2022direct}
Sun, C., Sun, M., Chen, H.: Direct voxel grid optimization: Super-fast convergence for radiance fields reconstruction. In: {CVPR} 2022. pp. 5449--5459 (2022)

\bibitem{tewari2019fml}
Tewari, A., Bernard, F., Garrido, P., Bharaj, G., Elgharib, M., Seidel, H., P{\'{e}}rez, P., Zollh{\"{o}}fer, M., Theobalt, C.: {FML:} face model learning from videos. In: {CVPR} 2019. pp. 10812--10822 (2019)

\bibitem{tewari2022advances}
Tewari, A., Thies, J., Mildenhall, B., Srinivasan, P.P., Tretschk, E., Wang, Y., Lassner, C., Sitzmann, V., Martin{-}Brualla, R., Lombardi, S., Simon, T., Theobalt, C., Nie{\ss}ner, M., Barron, J.T., Wetzstein, G., Zollh{\"{o}}fer, M., Golyanik, V.: Advances in neural rendering. Comput. Graph. Forum  \textbf{41}(2),  703--735 (2022)

\bibitem{thies2020neural}
Thies, J., Elgharib, M., Tewari, A., Theobalt, C., Nie{\ss}ner, M.: Neural voice puppetry: Audio-driven facial reenactment. In: {ECCV} 2020. vol. 12361, pp. 716--731 (2020)

\bibitem{thies2015real}
Thies, J., Zollh{\"{o}}fer, M., Nie{\ss}ner, M., Valgaerts, L., Stamminger, M., Theobalt, C.: Real-time expression transfer for facial reenactment. {ACM} Trans. Graph.  \textbf{34}(6),  183:1--183:14 (2015)

\bibitem{thies2016face}
Thies, J., Zollh{\"{o}}fer, M., Stamminger, M., Theobalt, C., Nie{\ss}ner, M.: Face2face: Real-time face capture and reenactment of {RGB} videos. In: {CVPR} 2016. pp. 2387--2395 (2016)

\bibitem{vaswani2017attention}
Vaswani, A., Shazeer, N., Parmar, N., Uszkoreit, J., Jones, L., Gomez, A.N., Kaiser, {\L}., Polosukhin, I.: Attention is all you need. In: NeurIPS (2017)

\bibitem{wang2022r2l}
Wang, H., Ren, J., Huang, Z., Olszewski, K., Chai, M., Fu, Y., Tulyakov, S.: R2l: Distilling neural radiance field to neural light field for efficient novel view synthesis. In: ECCV (2022)

\bibitem{wang2004image}
Wang, Z., Bovik, A.C., Sheikh, H.R., Simoncelli, E.P.: Image quality assessment: from error visibility to structural similarity. TIP  \textbf{13}(4),  600--612 (2004)

\bibitem{weise2011realtime}
Weise, T., Bouaziz, S., Li, H., Pauly, M.: Realtime performance-based facial animation. {ACM} Trans. Graph.  \textbf{30}(4), ~77 (2011)

\bibitem{weng2022humannerf}
Weng, C.Y., Curless, B., Srinivasan, P.P., Barron, J.T., Kemelmacher-Shlizerman, I.: Humannerf: Free-viewpoint rendering of moving people from monocular video. In: CVPR (2022)

\bibitem{yang2020facescape}
Yang, H., Zhu, H., Wang, Y., Huang, M., Shen, Q., Yang, R., Cao, X.: Facescape: {A} large-scale high quality 3d face dataset and detailed riggable 3d face prediction. In: {CVPR} 2020. pp. 598--607 (2020)

\bibitem{yang2019deep}
Yang, W., Zhang, X., Tian, Y., Wang, W., Xue, J.H., Liao, Q.: Deep learning for single image super-resolution: A brief review. TMM  \textbf{21}(12),  3106--3121 (2019)

\bibitem{yu2021plenoctrees}
Yu, A., Li, R., Tancik, M., Li, H., Ng, R., Kanazawa, A.: Plenoctrees for real-time rendering of neural radiance fields. In: ICCV (2021)

\bibitem{yu2023dylin}
Yu, H., Julin, J., Milacski, Z.A., Niinuma, K., Jeni, L.A.: Dylin: Making light field networks dynamic. In: CVPR (2023)

\bibitem{zhang2018unreasonable}
Zhang, R., Isola, P., Efros, A.A., Shechtman, E., Wang, O.: The unreasonable effectiveness of deep features as a perceptual metric. In: CVPR (2018)

\bibitem{zhang2018image}
Zhang, Y., Li, K., Li, K., Wang, L., Zhong, B., Fu, Y.: Image super-resolution using very deep residual channel attention networks. In: ECCV (2018)

\bibitem{zhao2022thin}
Zhao, J., Zhang, H.: Thin-plate spline motion model for image animation. In: CVPR (2022)

\bibitem{zheng2022avatar}
Zheng, Y., Abrevaya, V.F., B{\"u}hler, M.C., Chen, X., Black, M.J., Hilliges, O.: Im avatar: Implicit morphable head avatars from videos. In: CVPR (2022)

\bibitem{zheng2023pointavatar}
Zheng, Y., Yifan, W., Wetzstein, G., Black, M.J., Hilliges, O.: Pointavatar: Deformable point-based head avatars from videos. In: {CVPR} 2023. pp. 21057--21067. {IEEE} (2023)

\bibitem{zielonka2023instant}
Zielonka, W., Bolkart, T., Thies, J.: Instant volumetric head avatars. In: {CVPR} 2023. pp. 4574--4584 (2023)

\bibitem{zollhfer2018state}
Zollh{\"{o}}fer, M., Thies, J., Garrido, P., Bradley, D., Beeler, T., P{\'{e}}rez, P., Stamminger, M., Nie{\ss}ner, M., Theobalt, C.: State of the art on monocular 3d face reconstruction, tracking, and applications. Comput. Graph. Forum  \textbf{37}(2),  523--550 (2018)

\end{thebibliography}
